\PassOptionsToPackage{numbers,sort&compress}{natbib}
\documentclass{article}
\usepackage{hyphenat}

\usepackage[eandd, final]{neurips_2026}
\usepackage[utf8]{inputenc} 
\usepackage[T1]{fontenc}    
\usepackage{hyperref}       
\usepackage{url}            
\usepackage{booktabs}       
\usepackage{amsfonts}       
\usepackage{nicefrac}       
\usepackage{microtype}      
\usepackage{xcolor}         
\usepackage{amsmath,amssymb}
\usepackage{multirow}
\usepackage{graphicx}
\usepackage[numbers,sort&compress]{natbib}
\usepackage{enumitem}
\usepackage{xspace}
\usepackage{algorithm}
\usepackage{algpseudocode}
\usepackage{tabularx}
\usepackage{array}
\usepackage{makecell}
\definecolor{gray80}{gray}{0}
\usepackage{booktabs}
\usepackage{graphicx}
\usepackage{pifont}

\newif\ifcrhl
\crhlfalse
\newcommand{\crcolor}{\ifcrhl\color{blue}\fi}
\newcommand{\cradd}[1]{{\crcolor #1}}
\newenvironment{crblock}{\par\crcolor}{\par}
\newcommand{\cmark}{\ding{51}}
\newcommand{\xmark}{\ding{55}}

\newcommand{\best}[1]{\textbf{#1}}
\newcommand{\ours}{CSI-TextBench\xspace}

\title{CSI-TextBench: A Dataset and Benchmark for Language-Grounded Ambient Sensing Perception}

\author{%
  \textbf{Guozhen Zhu}$^{1}$\quad
  \textbf{Yuqian Hu}$^{1}$\quad
  \textbf{Sakila S Jayaweera}$^{1}$\quad
  \textbf{Wei-Hsiang Wang}$^{1}$\\
  \textbf{Beibei Wang}$^{1}$\quad
  \textbf{K. J. Ray Liu}$^{1}$\\[0.45em]
  $^{1}$ADT Research}

\begin{document}

\maketitle

\begin{abstract}
Indoor context deciphered by ambient sensing underpins many applications, from assisted living to smart buildings, yet remains weakly connected to the language-aligned paradigm of modern multimodal systems. Through billions of IoT devices, ambient WiFi Channel State Information (CSI) captures the indoor context of human activity and environmental dynamics passively via ubiquitous, always-on infrastructure. However, unlike vision or audio, CSI lacks naturally paired text like captions or transcripts, limiting its integration with language-driven AI systems.

We present CSI-TextBench, a large-scale dataset and benchmark for aligning CSI data with natural language, unifying 3 sensing benchmarks into 285,790 samples across 7 tasks and pairing them with 2,291 hierarchically structured, label-derived descriptions via many-to-many assignment to construct 2.0 million CSI–text training tuples. Benchmarking 5 alignment methods from 4 paradigm families, we show that contrastive CSI--text alignment matches supervised classifiers without class labels, adapts to unseen classes with few samples, and responds to open-vocabulary natural-language prompts. A single unified model supports compositional scene understanding, answering joint queries over activity, proximity, and identity, achieving 86.9\% top-1 accuracy on 140 composite scenes, outperforming task-specific classifier ensembles. CSI-TextBench enables ubiquitous ambient sensing as a language-indexed perception layer, enabling indoor contextual awareness in language-driven AI systems. \textit{Links: }\href{https://kaggle.com/datasets/b43cfe263881c7804a6b840820452782b4a0db647abdc94adcc982fdc5a72aab}{\textcolor{blue}{\textit{CSI-TextBench Dataset}}};
\href{https://github.com/guozhen-jenn-zhu/CSI-Text-Multimodal-Alignment.git}{\textcolor{blue}{\textit{CSI-TextBench Code}}}; 
\href{https://ai-csi-sensing.github.io/csi-text-bench/index.html}{\textcolor{blue}{\textit{Project Page}}}


\end{abstract}

\section{Introduction}
\label{sec:intro}

People spend a substantial portion of their daily lives indoors, creating a strong opportunity for AI systems to understand and support human activities through indoor contextual awareness. However, enabling continuous and fine-grained indoor perception remains an open challenge~\citep{du2024survey}. Among available sensing modalities, WiFi offers a particularly practical and scalable foundation: it is already ubiquitously deployed, always on, and inherently sensitive to human motion through Channel State Information (CSI). As people move, breathe, gesture, or occupy different locations, they perturb wireless signal propagation in ways that implicitly encode activity, proximity, and identity. This makes CSI a compelling substrate for passive and device-free ambient intelligence using existing infrastructure~\citep{9855454, wifiten}.

Ambient sensing via WiFi has made strong progress across activity recognition \cite{activity_3}, localization \cite{localization_1}, fall detection \cite{fall_2}, respiration monitoring \cite{respiration_1}, and user identification \cite{identification_1}, but most systems remain tied to fixed label spaces and predefined prediction targets, limiting how CSI can be queried, reused, and composed as a general-purpose perception interface \cite{WiFisurveyWang}. Language alignment offers a path forward: embedding signals and text in a shared space has transformed vision and audio into flexible, queryable systems with compositional understanding~\citep{radford2021clip,girdhar2023imagebind,akbari2021vatt}. For ambient sensing, however, it is blocked by data availability. Unlike images or speech, which benefit from co-occurring text like captions and transcripts, existing CSI datasets provide only coarse categorical labels, lacking the semantic richness and compositional structure required for signal--language alignment.



To address this gap, we introduce CSI-TextBench, a large-scale dataset and benchmark for aligning WiFi CSI with natural language. It turns existing CSI datasets into a unified signal--text resource. Built on this resource, the benchmark evaluates whether text-aligned CSI representations can move beyond fixed-label recognition to support zero-shot classification, few-shot transfer, open-vocabulary prompting, and compositional scene understanding. Our results show that CSI--language alignment can approach supervised classifiers under matched encoder architectures, adapt to unseen classes with few examples, and answer joint natural-language queries over activity, proximity, and identity from a shared embedding space. In the compositional setting, a single model reaches $86.9\%$ top-1 accuracy over 140 composite scenes.

CSI-TextBench makes three key contributions: (1) \textbf{a large-scale CSI--language resource} built from three public CSI benchmarks, which covers 285,790 samples across 7 tasks and 2,291 label-derived natural-language descriptions, yielding approximately 2.0M many-to-many CSI--text training tuples; (2)\textbf{ a controlled benchmark} of 5 alignment methods, revealing how objective design, text construction, and class granularity shape alignment quality; and (3) \textbf{standardized evaluation protocols} for zero-shot recognition, few-shot transfer, open-vocabulary prompting, and ambient question answering. 

CSI-TextBench aims to turn ambient sensing signals into a perception interface for indoor contextual awareness. By releasing the dataset, code, and benchmark, it provides a foundation for reusable, queryable, and language-compatible ambient intelligence on widely available wireless infrastructure.

\begin{figure*}[t]
    \centering
    \includegraphics[width=\textwidth]{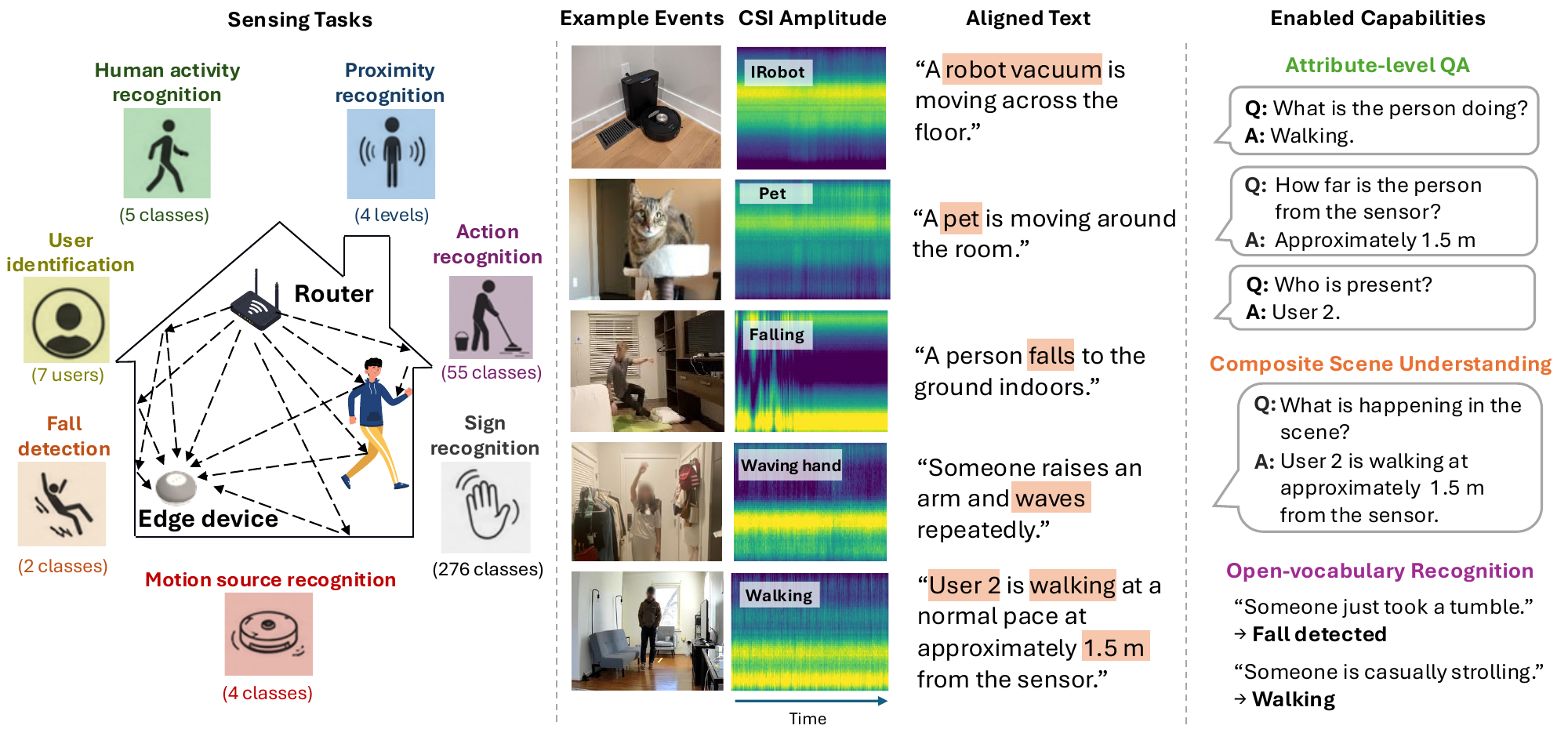}
    \caption{
    \textbf{CSI-TextBench overview.}
    CSI-TextBench pairs WiFi CSI recordings with natural-language descriptions across seven ambient sensing tasks. The benchmark supports language-grounded sensing, including QA, scene understanding, and open-vocabulary recognition.
    }
    \label{fig:concept}
\end{figure*}

\section{Related Work}
\label{sec:related}

\subsection{WiFi sensing}
WiFi CSI has been widely used for activity recognition~\citep{activity_3}, gesture and sign-language recognition~\cradd{\citep{ma2019signfi, wang2024xrf55, zheng2019widar3}}, fall detection \cradd{\cite{fall_2, regani2024fallaware}}, \cradd{localization~\citep{localization_1, wang2024roomlevel}, }proximity estimation \cite{proximity_1}, \cradd{motion source recognition~\citep{zhu2024srcsense}, }and user identification~\citep{identification_1}. Public benchmarks such as CSI-Bench~\citep{zhu2025csi}, XRF55~\citep{wang2024xrf55}, and SignFi~\citep{ma2019signfi} span diverse devices, environments, and semantic granularities, from coarse activity labels to 276 sign gestures. However, they provide categorical annotations rather than natural-language descriptions, leaving WiFi sensing largely disconnected from language-grounded multimodal learning.

\subsection{Multimodal alignment}
Language alignment has made vision, audio, and other modalities queryable through shared embedding spaces. CLIP~\citep{radford2021clip} showed that contrastive image--text alignment enables zero-shot transfer and compositional recognition. Subsequent work has explored sigmoid losses~\citep{zhai2023siglip}, anchor-based alignment~\citep{girdhar2023imagebind}, cross-correlation regularization~\citep{zbontar2021barlow}, and masked cross-modal prediction~\citep{singh2022flava}, with extensions to audio~\citep{guzhov2022audioclip, wu2023clap} and 3D point clouds~\citep{xue2023ulip}. These methods are largely developed for modalities with naturally co-occurring text and relatively direct signal--text pairing, whereas CSI requires label-derived language and induces dense many-to-many correspondence.

Recent RF-language studies show the promise of semantic wireless or radar sensing, including WiFi action captioning~\citep{wei2026wifi2cap}, large language model (LLM)-assisted WiFi recognition~\citep{zhang2025wichat}, RF--text contrastive gesture recognition~\citep{zhang2026wiclip}, mmWave zero-shot human action recognition (HAR)~\citep{cao2024mmclip}, radar-language motion understanding~\citep{lai2026radarllm}, and radar-based sign recognition~\citep{mineo2025trace}. Related efforts have also begun to align language with other sensing modalities, such as IMU-based activity reasoning~\citep{shouborno2025llasa}, and EEG-language modeling for neural decoding and clinical phenotyping~\citep{gijsen2025eeglanguage,mishra2025thought2text}. As summarized in Table~\ref{tab:sensing_language_datasets_benchmarks}, these works are typically tied to a single task, modality, or application setting, or do not release reusable sample--text pairs. CSI-TextBench complements them by providing an open, multi-task WiFi CSI--text dataset and benchmark for systematically evaluating RF--language alignment across diverse sensing tasks, semantic granularities, and alignment objectives.

\begin{table}[t]
\centering
\caption{Sensing-language datasets/benchmarks. N/A = not reported.}
\label{tab:sensing_language_datasets_benchmarks}
\resizebox{\textwidth}{!}{%
\begin{tabular}{llccclll}
    \toprule
    \textbf{Work} & \textbf{Signal} & \textbf{Open Source} & \textbf{Alignment} & \textbf{\#Task} & \textbf{\#Class} & \textbf{\#Sample} & \textbf{\#Pair} \\
    \midrule
    \textbf{WiFi2Cap~\citep{wei2026wifi2cap}} & WiFi CSI & \xmark & \cmark & 1 & 100 & N/A. & N/A. \\
    \textbf{Wi-Chat~\citep{zhang2025wichat}} & WiFi CSI & \xmark & \xmark & 1 & 4 & 393 & 393 \\
    \textbf{Wi-CLIP~\citep{zhang2026wiclip}} & WiFi CSI & \xmark & \cmark & 1 & 6 & N/A & N/A \\
    \midrule
    \textbf{mmCLIP~\citep{cao2024mmclip}} & mmWave radar & \cmark & \cmark & 1 & $>$200 & N/A & N/A\\
    \textbf{RadarLLM~\citep{lai2026radarllm}} & mmWave radar & \xmark & \cmark & 1 & N/A & 13,308 & 13,308\\
    \textbf{TRACE~\citep{mineo2025trace}} & FMCW radar & \xmark & \cmark & 1 & 126 & 25,830 & 25,830 \\
    \midrule
    \textbf{LLaSA~\citep{shouborno2025llasa}} & IMU & \cmark & \cmark & 3 & N/A & 188,429 & 235,661 \\
    \textbf{Sensor2Text~\citep{chen2024sensor2text}} & Wearable sensors & \xmark & \cmark & 1 & 20 & 17,100 & $\sim$7,700 \\
    \midrule
    \textbf{ELM-MIL~\citep{gijsen2025eeglanguage}} & EEG & \xmark & \cmark & 4 & 2--6 & 15,144 & 11,785 \\
    \textbf{Thought2Text~\citep{mishra2025thought2text}} & EEG & \cmark & \cmark & 1 & 40 & 11,940 & 11,940 \\
    \midrule
    \textbf{CSI-TextBench} & WiFi CSI & \cmark & \cmark & \textbf{7} & \textbf{353} & \textbf{285,790} & \textbf{2.0M} \\
    \bottomrule
\end{tabular}%
}
\end{table}


\section{The \ours Dataset}
\label{sec:dataset}

We introduce \ours, a large-scale dataset that pairs WiFi CSI with natural-language descriptions. It is designed to support systematic study of multimodal alignment for a sensing modality that does not naturally come with text.

We build a unified pipeline that converts heterogeneous CSI datasets into a common signal--text format. \ours integrates three public benchmarks, standardizes their CSI representations, and generates structured text descriptions from the original labels. In total, it covers 285,790 CSI samples across 7 tasks, paired with 2,291 unique descriptions through many-to-many assignment, yielding approximately 2.0M training tuples (Table~\ref{tab:dataset-stats}).

\paragraph{Source datasets.}
We combine three complementary CSI datasets that cover different sensing settings and levels of semantic granularity. \textbf{CSI-Bench}~\citep{zhu2025csi} provides large-scale in-the-wild data from commercial devices; we use five tasks from it: HAR, fall detection, motion source recognition, proximity, and user identification. \textbf{XRF55}~\citep{wang2024xrf55} contains 55 fine-grained indoor activities collected in controlled environments. \textbf{SignFi}~\citep{ma2019signfi} provides 276 sign-language gesture classes, making it the most fine-grained task in our benchmark.

We follow the licenses and release constraints of the source datasets. For SignFi, we release only our preprocessing scripts and CSI--text registry files due to restrictions; users obtain the original CSI data from the dataset providers. Together, these datasets span 2--276 classes, controlled and real-world settings, and diverse hardware configurations.

\paragraph{Unified preprocessing.}
The source datasets differ substantially in subcarrier counts, antenna configurations, and sampling rates. To ensure consistency, we follow the CSI-Bench preprocessing pipeline~\citep{zhu2025csi} and standardize all inputs into a unified amplitude tensor $\mathbf{X} \in \mathbb{R}^{C \times K \times T}$, where $C$ denotes the number of channels (antenna pairs), $K$ the number of aligned subcarriers (frequency bins), and $T$ the temporal length of each sample. The preprocessing includes amplitude extraction, per-sample normalization, and subcarrier alignment to a fixed $K$. 

\paragraph{Text construction.}
CSI signals do not have native text annotations, so we derive descriptions from task labels using hand-written templates. For each class, we create multiple semantically equivalent descriptions to increase linguistic diversity while preserving label semantics.

When multiple labels are available, we use two caption levels. Level~1 captions describe a single attribute, such as activity or gesture. Level~2 captions combine multiple attributes into a scene-level description, e.g., \textit{``User~3 is walking at approximately 1.5\,m from the sensor''}. This supports evaluation of both single-attribute alignment and multi-attribute scene understanding.

All descriptions are restricted to information observable from CSI, including activity, gesture, proximity, motion source, and user-related behavioral patterns. We exclude deployment-specific details such as device type and room identity to reduce dataset-specific shortcuts. We use hand-written rather than LLM-generated templates to ensure label grounding, controlled evaluation, and reproducibility. Generalization to novel language is evaluated separately with unseen prompts (\S\ref{sec:exp-generalization}).

\paragraph{Many-to-many structure.}
\label{sec:m2m}
Each CSI sample is paired with 4--9 descriptions, and each description can match many CSI samples. The resulting benchmark is explicitly many-to-many: different signals may express the same semantics, and different texts may describe the same event. Unlike standard one-to-one image--text pairing, this introduces ambiguity in positive and negative pairs, making it a useful testbed for alignment under many-to-many correspondence.

Overall, \ours provides a unified large-scale benchmark for CSI--language alignment across heterogeneous sensing conditions, diverse semantic scopes, and dense many-to-many signal--text relationships.

\begin{table*}[t]
\centering
\caption{Dataset statistics for \ours. The three multitask CSI-Bench tasks (HAR, Proximity, User~ID) share the same 55,684 underlying CSI recordings; each appears separately. The 285,790 task-level instances correspond to 174,422 unique recordings.}
\label{tab:dataset-stats}
\small
\begin{tabular}{llrrrrr}
\toprule
Task & Dataset & Classes & CSI Samples & Unique Texts & Texts/Sample & Total Pairs \\
\midrule
HAR              & CSI-Bench & 5   & 55,684  & 221   & 9 & 501,156 \\
Proximity        & CSI-Bench & 4   & 55,684  & 216   & 9 & 501,156 \\
User ID          & CSI-Bench & 7   & 55,684  & 224   & 8 & 445,472 \\
Fall Detection   & CSI-Bench & 2   & 6,700   & 10    & 5 & 33,500 \\
Motion Source    & CSI-Bench & 4   & 60,858  & 20    & 5 & 304,290 \\
Sign Recognition & SignFi    & 276 & 8,280   & 1,380 & 5 & 41,400 \\
Action Recognition & XRF55   & 55  & 42,900  & 220   & 4 & 171,600 \\
\midrule
\textbf{Total}   &           & \textbf{353} & \textbf{285,790} & \textbf{2,291} & & \textbf{1,998,574} \\
\bottomrule
\end{tabular}
\end{table*}

\section{Benchmark Alignment Methods}
\label{sec:methods}
 
We benchmark representative alignment methods under a unified architecture to establish strong baselines for CSI--text learning. CSI--text alignment differs from more standard multimodal settings in several important respects: correspondence is highly many-to-many, large-scale naturally paired data is unavailable, and the unique signal characteristics of CSI as an underexplored sensing modality. These factors motivate a controlled comparison of diverse alignment paradigms under a shared benchmark.
 
\subsection{Shared Architecture}
\label{sec:arch}
 
All methods share a common dual-encoder design to isolate the effect of the alignment objective. 
A CSI encoder $f_\theta$ processes input $\mathbf{x} \in \mathbb{R}^{C \times K \times T}$, followed by a projection head $h_\psi$ to produce embeddings $z_{\text{csi}} \in \mathbb{R}^d$. 
Text is encoded using a frozen pretrained encoder $g_\phi$ with projection head $h_\omega$, yielding $z_{\text{text}} \in \mathbb{R}^d$. 
Both embeddings are $\ell_2$-normalized before computing alignment losses, with default dimension $d=256$.
 
The text encoder is kept frozen to preserve its pretrained semantic structure, ensuring that learning primarily occurs in the CSI encoder. This setup isolates cross-modal alignment from language modeling effects. Encoder choices are analyzed in \S\ref{sec:exp-ablations}.

\subsection{Alignment Objectives}
\label{sec:alignment_methods}

 We evaluate 5 methods spanning four paradigm families to probe different assumptions about cross-modal alignment under dense many-to-many correspondence. Full formulations are in Appendix~\ref{app:methods}.

\textbf{Contrastive:} CLIP-style InfoNCE~\citep{radford2021clip} uses a softmax-normalized batch contrastive objective that implicitly treats only the paired item as positive within each batch, creating false-negative pressure when multiple CSI samples or texts share the same semantics. SigLIP~\citep{zhai2023siglip} replaces softmax normalization with independent pairwise sigmoid losses, which can accommodate multiple positives per anchor when the positive-pair matrix is defined accordingly.

\textbf{Anchor-based:} We adapt LanguageBind~\citep{zhu2024languagebind} by aligning the CSI encoder to a frozen text embedding space with an asymmetric contrastive objective. This tests whether leveraging pretrained language structure can compensate for limited paired CSI--text data.

\textbf{Non-contrastive:} We adapt Barlow Twins~\citep{zbontar2021barlow} to cross-modal alignment by enforcing cross-correlation consistency between embedding dimensions, avoiding explicit negative sampling---a useful property when in-batch negatives may share semantics with positives.

\textbf{Reconstructive:} A masked cross-modal prediction objective (Masked Pred.), inspired by masked multimodal modeling~\citep{singh2022flava,bachmann2022multimae}, reconstructs the text embedding from a partially masked CSI input, testing whether reconstructive supervision captures structure beyond purely discriminative objectives.
 
\section{Experiments}
\label{sec:experiments}
 
We evaluate along three progressive layers, each posed as a distinct question. \textbf{Layer~1} in \S\ref{sec:exp-alignment} asks whether text-aligned CSI encoders, trained without class labels, can match fully supervised classifiers. \textbf{Layer~2} in \S\ref{sec:exp-generalization} asks whether the resulting embeddings generalize beyond the training distribution along two axes: unseen classes and unseen language. \textbf{Layer~3} in \S\ref{sec:exp-ambient} asks whether a single text-aligned model can serve as a perceptual backbone for ambient intelligence, providing a capability that no collection of independent supervised classifiers can structurally replicate. Design ablations follow in \S\ref{sec:exp-ablations}.
 
\subsection{Experimental Setup}
\label{sec:exp-setup}
 
\paragraph{Architecture.}
The CSI encoder is a 1D-adapted ResNet-18 \cite{kaiming2016}: input reshaped to $[B, C{\cdot}K, T]$, a Conv1d stem with kernel size $7$ and stride $2$, four residual stages with channel widths $64{\to}128{\to}256{\to}512$, adaptive average pooling, and linear projection to dimension $256$. The text encoder is a frozen CLIP ViT-B/32 with output dimension $512$. Each modality has a 2-layer MLP projection mapping to a shared 256-dimensional space, with BatchNorm, GELU, and L2 normalization. Alternatives are compared in \S\ref{sec:exp-ablations}.
 
\paragraph{Training.}
We train 35 single-task models spanning 5 alignment methods and 7 tasks (\S\ref{sec:methods}). All models use AdamW~\cite{Adam19} with learning rate $1{\times}10^{-3}$, weight decay $1{\times}10^{-5}$, cosine annealing with 5-epoch warmup, gradient clipping at $1.0$, batch size $128$, and early stopping with patience 15. 
The supervised baseline uses the same encoder with a linear head trained by cross-entropy with label smoothing 0.1, providing a matched-capacity reference. Appendix~\ref{app:sota-context} compares this baseline with published results from each source dataset. Training runs on AWS SageMaker ml.g5.xlarge instances (NVIDIA A10G, 24GB), requiring 15--45 minutes per single-task model and 2--4 hours for multitask training.
 
\paragraph{Evaluation.}
Zero-shot classification assigns the class whose mean text-prototype embedding has highest cosine similarity to the test CSI embedding. Retrieval ranks prototypes by cosine similarity. Few-shot linear probing fits logistic regression with L-BFGS and $C{=}1$ on $k$ frozen embeddings per class, averaged over 10 trials.

\subsection{Layer 1: Alignment Effectiveness}
\label{sec:exp-alignment}
 
\paragraph{Zero-shot classification.}
We train 5 CSI--text alignment methods on seven tasks, yielding 35 models, and evaluate them with zero-shot prototype matching against a supervised classifier using the same encoder (Table~\ref{tab:zeroshot}).
CLIP InfoNCE performs best, coming within $0.3$ points of the supervised classifier on HAR, motion source, and user ID, and within $2.4$ points on the 276-class SignFi task. 4 of the 5 methods exceed $92\%$ on the simpler tasks, showing that CSI--text alignment is generally achievable. Unlike the supervised baseline, it does not train a task-specific classification head or use class labels as direct categorical targets; instead, supervision comes through label-derived natural-language descriptions. 
Sigmoid and asymmetric losses are unstable on SignFi, while Barlow Twins collapses on fall detection, motion source, and SignFi. These results suggest that label-free text-aligned encoders can approach supervised performance, but only when the objective explicitly contrasts CSI samples against text descriptions.

\begin{table}[t]
\centering
\caption{Zero-shot classification accuracy (\%, mean$\pm$std over 3 seeds).}
\label{tab:zeroshot}
\vspace{2pt}
\setlength{\tabcolsep}{3.2pt}
\small
\begin{tabular}{@{}lccccccc@{}}
\toprule
\textbf{Method} & \textbf{HAR} & \textbf{Proximity} & \textbf{User ID} & \textbf{Fall} & \textbf{MotSrc} & \textbf{SignFi} & \textbf{XRF55} \\
\midrule
CLIP InfoNCE & \best{97.8}{\scriptsize$\pm$0.6} & \best{94.6}{\scriptsize$\pm$0.4} & \best{100.0}{\scriptsize$\pm$0.0} & 94.2{\scriptsize$\pm$0.0} & \best{99.5}{\scriptsize$\pm$0.2} & \best{97.0}{\scriptsize$\pm$1.0} & 77.5{\scriptsize$\pm$0.4} \\
SigLIP & 95.7{\scriptsize$\pm$2.8} & 84.4{\scriptsize$\pm$3.8} & 100.0{\scriptsize$\pm$0.0}  & \best{95.5}{\scriptsize$\pm$0.5}  & 99.5{\scriptsize$\pm$0.3} & 71.8{\scriptsize$\pm$32.2} & 74.5{\scriptsize$\pm$1.2} \\
LanguageBind & 95.1{\scriptsize$\pm$2.8} & 76.0{\scriptsize$\pm$3.3} & 100.0{\scriptsize$\pm$0.0}  & 95.3{\scriptsize$\pm$0.3} & 99.4{\scriptsize$\pm$0.1} & 75.7{\scriptsize$\pm$29.8} & 77.6{\scriptsize$\pm$0.4} \\
Barlow Twins & 92.6{\scriptsize$\pm$1.1} & 88.8{\scriptsize$\pm$5.6} & 98.8{\scriptsize$\pm$1.3} & 59.8{\scriptsize$\pm$6.8} & 63.2{\scriptsize$\pm$1.4} & 61.3{\scriptsize$\pm$27.4} & 65.1{\scriptsize$\pm$3.8} \\
Masked Pred. & 94.2{\scriptsize$\pm$2.8} & 92.1{\scriptsize$\pm$2.7} & 99.2{\scriptsize$\pm$0.6} & 94.1{\scriptsize$\pm$0.6} & 96.1{\scriptsize$\pm$1.8} & 95.7{\scriptsize$\pm$6.0} & \best{78.3}{\scriptsize$\pm$2.0} \\
\midrule
\emph{Supervised} & 97.9 & 96.3 & 100.0 & 95.5 & 99.8 & 99.4 & 82.9 \\
\bottomrule
\end{tabular}
\end{table}

\paragraph{Cross-modal retrieval.}
Top-1 accuracy only checks whether the correct prototype is ranked first, but it does not reveal whether the full embedding space is well structured. This matters for open-set and compositional tasks, where the model must compare many possible text queries. We therefore evaluate cross-modal retrieval (Table~\ref{tab:retrieval}).
CLIP InfoNCE performs strongly on the large-class tasks, reaching R@5 of $95.4\%$/$99.1\%$ and R@10 of $98.0\%$/$99.7\%$ on XRF55/SignFi. Other methods fall to $57$--$62\%$ R@5 on SignFi, indicating that their embeddings are less globally coherent. Thus, CLIP InfoNCE not only produces correct top-1 predictions, but also organizes the CSI--text space more reliably, enabling the open-set evaluations that follow. \cradd{Appendix~\ref{app:m2m-retrieval} complements Recall@$K$ with mAP and R-Precision over the full HAR caption gallery under strict and class-level relevance.}

\begin{table}[t]
\centering
\caption{Cross-modal retrieval on R@1(\%). R@5 and R@10 for large-class tasks.}
\label{tab:retrieval}
\vspace{2pt}
\small
\resizebox{\textwidth}{!}{%
\begin{tabular}{@{}lcccccccccc@{}}
\toprule
 & \multicolumn{7}{c}{\textbf{CSI$\to$Text R@1}} & \multicolumn{2}{c}{\textbf{R@5}} & \multicolumn{1}{c}{\textbf{R@10}} \\
\cmidrule(lr){2-8} \cmidrule(lr){9-10} \cmidrule(l){11-11}
\textbf{Method} & HAR & Proximity & User ID & Fall & MotSrc & SignFi & XRF55 & SignFi & XRF55 & SignFi \\
\midrule
CLIP InfoNCE & 90.2 & \best{94.9} & 99.9 & 94.2 & \best{48.3} & \best{95.8} & \best{78.0} & \best{99.1} & \best{95.4} & \best{99.7} \\
SigLIP & 91.9 & 84.7 & \textbf{100.0} & \best{95.9} & 48.0 & 26.4 & 73.0 & 57.9 & 93.9 & 72.3 \\
LanguageBind & 91.2 & 79.6 & 100.0 & 95.6 & 48.3 & 33.6 & 77.6 & 57.5 & 94.0 & 69.5 \\
Barlow Twins & 94.1 & 80.9 & 99.8 & 61.6 & 20.3 & 22.7 & 66.4 & 55.3 & 81.4 & 71.0 \\
Masked Pred. & \best{94.6} & 95.5 & 98.9 & 93.7 & 47.6 & 87.3 & 75.4 & 97.7 & 93.2 & 99.6 \\
\bottomrule
\end{tabular}%
}
\end{table}

\subsection{Layer 2: Generalization}
\label{sec:exp-generalization}
 
Supervised WiFi sensing models are closed-set: they can only predict classes seen during training and cannot directly respond to new text queries at inference time. We therefore test whether a text-aligned encoder generalizes in two ways: to unseen classes and to unseen language.

\paragraph{Unseen-class generalization.}
We train CLIP InfoNCE on 200 of the 276 SignFi classes and evaluate transfer to the remaining 76 unseen classes using zero-shot prototype matching, cross-modal retrieval, and few-shot linear probing on frozen embeddings (10 trials). Results are shown in Table~\ref{tab:holdout}. Zero-shot accuracy on unseen classes is $11.3\%$ ($8.5\times$ chance). While still well below the $96.4\%$ accuracy on seen classes, this is an unsurprising result given the fine-grained nature of the label space: many sign classes differ by only one or two words in text, limiting separability under text-only transfer. Retrieval results reveal a stronger underlying structure: the correct class has median rank 12 out of 76 (top $16\%$), and R@10 reaches $46.1\%$. Thus, even when top-1 zero-shot prediction fails, the correct class is often placed near the target in embedding space.

Adaptation to new classes is highly efficient. With only 1 labeled example per class, accuracy increases to $54.0\%$; with 10 examples, it reaches $74.6\%$. These results show that the encoder learns a transferable representation for unseen gestures, enabling rapid extension to new classes with a lightweight probe on frozen embeddings rather than full-model retraining.

\begin{table}[t]
\centering
\caption{Unseen-class generalization on SignFi. Retrieval metrics (R@$K$, Med.\ Rank) are on held-out classes only.}
\label{tab:holdout}
\small
\resizebox{\textwidth}{!}{%
\begin{tabular}{@{}lcccccccc@{}}
\toprule
 & \multicolumn{2}{c}{Zero-shot Accuracy} & \multicolumn{3}{c}{Retrieval (held-out)} & \multicolumn{3}{c}{Few-shot Accuracy} \\
\cmidrule(lr){2-3} \cmidrule(lr){4-6} \cmidrule(l){7-9}
 & Seen (200) & Held-out (76) & R@5 & R@10 & Med.\ Rank & 1-shot & 5-shot & 10-shot \\
\midrule
Performance (\%) & 96.4 & 11.3 & 29.7 & 46.1 & 12 & 54.0{\scriptsize$\pm$3.0} & 64.7{\scriptsize$\pm$1.5} & 74.6{\scriptsize$\pm$1.1} \\
\bottomrule
\end{tabular}
}
\end{table}
 
\begin{table}[t]
\centering
\caption{Open-vocabulary recognition accuracy on HAR (zero-shot).}
\label{tab:openvocab}
\vspace{2pt}
\small
\resizebox{\textwidth}{!}{%
\begin{tabular}{@{}lcccccc@{}}
\toprule
 & Train template & Layperson & Expert (raw) & Expert (anch.) & Sensor (raw) & Sensor (anch.) \\
\midrule
Accuracy (\%) & 95.9 & \best{96.0} & 59.1 & 96.2 & 35.4 & 77.1 \\
\bottomrule
\end{tabular}
}
\end{table}

\paragraph{Open-vocabulary recognition.}
We next test whether the model can handle new natural-language prompts at inference time, rather than only the templates used during training. This is important because strong performance on new prompts would indicate real text alignment, not memorization of fixed phrases. We replace the HAR text prototypes with unseen prompts from three registers: layperson language, e.g., \textit{``Someone just took a tumble''}; domain-expert language, e.g., \textit{``Anterior--posterior center-of-pressure shift consistent with gait initiation''}; and sensor-grounded language, e.g., \textit{``Quasi-periodic Doppler modulation from bipedal locomotion''}. For the two technical registers, we also test anchored variants that place the terminology in a more natural sentence form, closer to CLIP's pretraining distribution. Results are reported in Table~\ref{tab:openvocab}.

Layperson prompts achieve $96.0\%$ accuracy, matching the $95.9\%$ training-template baseline despite using different wording. Raw technical prompts perform much worse, with $59.1\%$ for domain-expert prompts and $35.4\%$ for sensor-grounded prompts. However, anchoring these prompts improves accuracy to $96.2\%$ and $77.1\%$, respectively. These results show that the model is not simply memorizing training templates: it can respond to new language when the prompts remain close to the text encoder's semantic space. At the same time, performance drops for more specialized language, suggesting that open-vocabulary coverage \cradd{depends on both the training language and the pretrained text encoder (\S\ref{sec:exp-freeform}).}

\begin{crblock}
\subsection{Human-Written Free-Form Language}
\label{sec:exp-freeform}
Section~\ref{sec:exp-generalization} replaces the training templates with unseen prompts at inference time. We further test alignment with independently written language through a human-written free-form layer for HAR. Human writers described the activities using the collection-session logs and activity descriptions as references, independently of the template registry and without LLM assistance, in five registers: conversational, neutral-descriptive, field-note, log-style, and hedged, e.g., \textit{``Someone's waving hello (or maybe goodbye, hard to tell)''}. We train a template-supervised and a free-form-supervised model with identical architecture, splits, and optimization, and evaluate both on the same prompt families, including held-out human-written prompts seen by neither model (Table~\ref{tab:freeform}; details in Appendix~\ref{app:freeform}).

\begin{table}[t]\crcolor
\centering
\caption{HAR zero-shot accuracy (\%) of template- and free-form-supervised models. Ens.\ averages five unseen prompts per class; held-out free-form is the mean over held-out human-written prompt families.}
\label{tab:freeform}
\vspace{2pt}
\small
\resizebox{\textwidth}{!}{%
\begin{tabular}{@{}lcccccc@{}}
\toprule
Training text & Train template & Layperson & Expert (raw) & Sensor (raw) & Sensor (ens.) & Held-out free-form \\
\midrule
Template & 95.9 & 96.0 & 59.1 & 35.4 & 76.4 & 95.0 \\
Free-form & 94.5 & 93.9 & 87.0 & 80.2 & 84.0 & 94.0 \\
\bottomrule
\end{tabular}
}
\end{table}

The template-trained model reaches $95.0\%$ on held-out free-form prompts, $0.9$ points below its training-template accuracy, so template supervision transfers to independently written everyday language. Free-form supervision raises raw domain-expert and sensor-grounded accuracy by $27.9$ and $44.8$ points, and lowers training-template, layperson, and held-out free-form accuracy by $1.0$--$2.1$ points. Broader human-written supervision therefore extends coverage of specialized language with a small cost on ordinary language.
\end{crblock}
 
\subsection{Layer 3: Ambient Scene Perception}
\label{sec:exp-ambient}


We next ask whether a single text-aligned model can serve as a queryable perception interface for ambient sensing. From one WiFi signal, the model should answer natural-language questions about who is present, what they are doing, and how far they are from the sensor. This is difficult for standard supervised classifiers, which are tied to fixed prediction targets and must be combined manually to describe a full scene. A text-aligned model offers a different approach: the natural-language query defines the task at inference time, enabling more flexible and compositional scene understanding.

\paragraph{Unified multitask recognition.}
 
We train a single CLIP-InfoNCE model jointly on all 7 tasks with proportional sampling and the same zero-shot protocol as in \S\ref{sec:exp-alignment}. Text prototypes are constructed independently for each task, and the CSI embedding is classified by argmax cosine similarity. The model uses one shared cross-dataset encoder with no task-specific heads or routing. The unified model preserves at least $95\%$ of the corresponding single-task accuracy on 6 of 7 tasks, including $96.0\%$ on HAR and $96.1\%$ on the 276-class SignFi task (Table~\ref{tab:multitask}). Only Proximity drops, from $94.6\%$ to $84.5\%$, likely from competition with HAR over shared spatial features. Overall, the results show that a single encoder can support heterogeneous tasks across datasets, with the task specified only through text prototypes. This makes the model task-agnostic and directly usable as a unified queryable perception layer.

\begin{table}[t]
\centering
\caption{Unified multitask model: zero-shot accuracy across 7 tasks, no task-specific heads. F1 within 1\,pp throughout.}
\label{tab:multitask}
\vspace{2pt}
\small
\begin{tabular}{@{}lccccccc@{}}
\toprule
 & HAR & Proximity & User ID & Fall & MotSrc & SignFi & XRF55 \\
\midrule
Accuracy (\%) & 96.0 & 84.5 & 100.0 & 95.9 & 98.2 & 96.1 & 79.2 \\
\bottomrule
\end{tabular}
\end{table}
 
\begin{table}[t]
\centering
\caption{Ambient scene QA and composite-scene retrieval accuracy (\%) on the co-labeled CSI-Bench subset.}
\label{tab:ambient-qa}
\vspace{2pt}
\small
\setlength{\tabcolsep}{4.5pt}
\begin{tabular}{@{}lccccccc@{}}
\toprule
 & HAR & Proximity & User ID & Joint / Top-1 & Top-3 & Top-5 & MRR \\
\midrule
L1: Per-attribute  & 94.6 & 82.3 & 88.3 & 70.4 & -- & -- & -- \\
L2: Composite      & 96.6 & 89.2 & 99.9 & \best{86.9} & 96.9 & 98.2 & 0.919 \\
\bottomrule
\end{tabular}
\end{table}

\paragraph{Compositional scene understanding.}
We next test whether the model can jointly infer multiple attributes of the same scene, rather than classify each attribute separately. We train a 3-task model on the co-labeled CSI-Bench subset, where each CSI recording has labels for HAR, proximity, and UserID.

\textbf{Level~1: per-attribute QA.}
For each attribute, we ask a natural-language question with candidate answers. For example, the question \textit{``What is the person doing?''} is paired with answers such as \textit{``Walking at a normal pace''} or \textit{``Sitting still and breathing calmly''}. Each question--answer pair is encoded as text, and the model selects the answer whose embedding is closest to the CSI embedding. Predictions are made independently for activity, proximity, and identity. Joint accuracy requires all three answers to be correct for the same sample.

\paragraph{Level~2: composite scene retrieval.}
We also evaluate a more holistic setting. We construct $140$ composite scene descriptions from 5 activities $\times$ 4 proximity levels $\times$ 7 users, such as \textit{``User~3 is walking at approximately 1.5\,m from the sensor''}. The model then retrieves the closest full-scene description in one step, jointly resolving all three attributes.

As summarized in Table~\ref{tab:ambient-qa}, composite retrieval achieves $86.9\%$ top-1 accuracy, outperforming the $70.4\%$ joint accuracy from independent per-attribute QA. This gap shows the limitation of factorizing the scene into separate predictions: multiplying the per-attribute accuracies gives an expected joint accuracy of ${\sim}68.7\%$, close to the observed $70.4\%$. In contrast, composite text embeddings can capture dependencies across activity, proximity, and identity. Level~2 also reaches $96.9\%$ top-3 accuracy and an MRR of $0.919$, showing that language-based composition provides a stronger way to recover the full scene state without adding task-specific heads or explicit compositional supervision. \cradd{Appendix~\ref{app:hoc} evaluates attribute combinations withheld from training.}

\subsection{Design Ablations}
\label{sec:exp-ablations}
 
We ablate three design choices to check whether the results depend on a specific model component. Unless otherwise noted, ablations use CLIP InfoNCE on HAR. Results are reported in Table~\ref{tab:ablations}.

\paragraph{CSI encoder.}
We compare ResNet-18 1D with ViT-1D and PatchTST~\cite{patchtst2023}. ResNet-18 1D performs best, reaching $98.7\%$ zero-shot accuracy, compared with $95.5\%$ for ViT-1D and $94.1\%$ for PatchTST. This suggests that convolutional structure is better suited to CSI at this data scale, likely because CSI contains local temporal patterns and shift-like variations.

\paragraph{Text encoder.}
We replace the frozen CLIP ViT-B/32 text encoder with frozen all-MiniLM-L6-v2 to test the role of text representation quality. Zero-shot accuracy drops from $98.7\%$ to $90.2\%$. With linear probing, the gap is smaller, from $99.0\%$ to $95.7\%$. This indicates that CLIP provides more useful text prototypes for zero-shot matching, while a learned probe can partially compensate for weaker text geometry.

\paragraph{Text granularity.}
We compare attribute-only Level~1 templates with richer Level~1+2 compositional templates. On HAR and Proximity, the difference is at most $0.3$ points, suggesting that compositional descriptions do not substantially improve single-task classification. Their main benefit appears in multitask ambient QA (\S\ref{sec:exp-ambient}), where the model must reason over multiple attributes jointly.

\begin{table}[t]
\centering
\caption{Design ablations. \underline{Underlined}: default. LP = linear probe on frozen embeddings.}
\label{tab:ablations}
\vspace{2pt}
\small
\begin{tabular}{@{}llccc@{}}
\toprule
\textbf{Ablation} & \textbf{Variant} & \textbf{Zero-shot Accuracy} & \textbf{F1} & \textbf{LP Accuracy} \\
\midrule
\multirow{3}{*}{\makecell[l]{CSI Encoder\\{\scriptsize (HAR)}}} & \underline{ResNet-18 1D} & \best{98.7\%} & \best{98.7\%} & \best{99.0\%} \\
 & ViT-1D & 95.5\% & 95.6\% & 96.3\% \\
 & PatchTST & 94.1\% & 94.2\% & 95.8\% \\
\midrule
\multirow{2}{*}{\makecell[l]{Text Encoder\\{\scriptsize (HAR)}}} & \underline{CLIP ViT-B/32} & \best{98.7\%} & \best{98.7\%} & \best{99.0\%} \\
 & MiniLM-L6-v2 & 90.2\% & 90.5\% & 95.7\% \\
\midrule
\multirow{2}{*}{\makecell[l]{Text Granularity\\{\scriptsize (HAR / Prox)}}} & \underline{Level 1} & \best{98.0\%} / 94.6\% & \best{98.1\%} / 94.7\% & \best{98.7\%} / -- \\
 & Level 1+2 & 97.7\% / \best{94.8\%} & 97.8\% / \best{94.9\%} & 98.3\% / -- \\
\bottomrule
\end{tabular}
\end{table}

 
\section{Limitation and Future Work}
\label{sec:limitations}
 
CSI-TextBench highlights a broader challenge in connecting physical sensing signals with language. While the current framework provides a controlled and signal-grounded setting for studying this problem, further progress will likely require advances beyond scaling data or adding evaluations. In particular, specialized or sensor-grounded language remains difficult for pretrained text encoders, and fine-grained signal distinctions may not always be fully expressed by template-based descriptions. \cradd{Open-vocabulary results concern unseen phrasing of concepts represented in training, and transfer to unseen classes remains limited on fine-grained tasks (\S\ref{sec:exp-generalization}). Cross-environment and cross-device performance remains constrained by the domain sensitivity of WiFi CSI (Appendix~\ref{app:cross-domain}).} These findings point to future opportunities in domain-aware text representations, alignment objectives tailored to sensing semantics, and model architectures that better bridge signal-level variation with language-level concepts. By providing a unified benchmark and evaluation suite, CSI-TextBench offers a foundation for developing these next-generation CSI--language models.
 
 
\section{Conclusion}
\label{sec:conclusion}
 
We presented CSI-TextBench, the first dataset and benchmark for aligning WiFi CSI with natural language. By unifying three sensing benchmarks into 285,790 samples across 7 tasks paired with hierarchically generated descriptions, we enable systematic study of CSI–language alignment. Across 5 methods from four paradigm families, language alignment matches supervised classifiers, transfers to unseen classes via few-shot adaptation, accepts open-vocabulary queries, and supports compositional scene understanding at 86.9\% over 140 joint configurations—capabilities the supervised paradigm cannot structurally provide.
We release the dataset, code, and benchmark to support research at the intersection of multimodal alignment and ambient sensing, toward ambient sensing AI systems that perceive indoor environments through the ubiquitous, always-on  WiFi devices.

\section{Broader Impact}
\label{sec:broader-impact}
 
WiFi CSI captures no identifiable visual or audio content, offering a privacy advantage over camera- and microphone-based monitoring and making language-grounded ambient sensing well-suited to assistive applications such as elderly care, fall detection, respiration monitoring, and smart-home automation on infrastructure already present in most indoor environments. By turning fixed-label sensors into queryable, composable perception layers, CSI-TextBench lowers the integration cost for downstream agents: a single shared encoder answers natural-language questions about who is present, what they are doing, and where, without retraining for each new query. The same recipe transfers directly to other ambient modalities lacking natural text—radar, mmWave, acoustic sensing—offering a general path to language-grounded ambient perception across the broader sensing community.
As with any technology that detects human presence and activity, responsible deployment matters; we encourage explicit consent mechanisms, on-device processing, and access controls for ambient sensing systems. \cradd{Out-of-scope uses include covert monitoring and identifying individuals without their consent.} CSI-TextBench builds on publicly available datasets collected under their respective IRB approvals. \cradd{Appendix~\ref{app:datasheet} summarizes intended uses and privacy considerations.}

\newpage

\bibliographystyle{plain}
\bibliography{references}
 
\newpage
\appendix
 
 
\section{Preprocessing Details}
\label{app:preprocessing}

The three source datasets are collected using different hardware platforms, so they differ in subcarrier count, antenna configuration, sampling rate, and CSI extraction firmware. To make them compatible, we apply the same preprocessing pipeline used in CSI-Bench~\citep{zhu2025csi} to all datasets.

\begin{enumerate}[leftmargin=*,itemsep=2pt]
\item \textbf{Amplitude extraction.}
Raw CSI measurements contain phase noise caused by device-level timing and frequency synchronization offsets. Following CSI-Bench~\citep{zhu2025csi}, we discard the phase and keep only the amplitude $|H(f_k, t)|$ at subcarrier $f_k$ and time step $t$.

\item \textbf{Data segmentation.}
We divide continuous recordings into non-overlapping 5-second windows, following the protocol of CSI-Bench.

\item \textbf{Amplitude normalization.}
Each sample is normalized independently to have zero mean and unit variance:
\[
\hat{H}(f_k, t) = \frac{H(f_k, t) - \mu_H}{\sigma_H + \epsilon}.
\]
This reduces absolute signal-strength differences across devices and environments.

\item \textbf{Subcarrier standardization.}
Different chipsets provide different numbers of subcarriers (e.g., 14 for Broadcom, 64 for ESP32, 256 for Qualcomm, and 270 for Intel 5300). We therefore standardize the frequency dimension to a fixed size $K$ using zero-padding or uniform subsampling, and obtain a unified tensor representation $\mathbf{X} \in \mathbb{R}^{C \times K \times T}$.
\end{enumerate}

After these steps, all datasets share the same amplitude-tensor format. This allows us to train one CSI encoder across datasets without changing the model architecture.
 
\section{Alignment Method Formulations}
\label{app:methods}

All methods use the same shared dual-encoder architecture and produce $\ell_2$-normalized embeddings $z_{\text{csi}}, z_{\text{text}} \in \mathbb{R}^d$ (Section~\ref{sec:arch}). Unless otherwise noted, we compute the batch similarity matrix as
\[
S_{ij} = \frac{z_{\text{csi}}^{i\top} z_{\text{text}}^j}{\tau},
\]
where $\tau$ is a learnable temperature, initialized to 0.07 and optimized in log-space.

We compare five alignment objectives that make different assumptions about how CSI and text should be matched.

\paragraph{CLIP-InfoNCE}~\citep{radford2021clip}.
This is a symmetric contrastive loss. It treats each matched CSI-text pair on the diagonal of the similarity matrix as a positive pair:
\begin{equation}
  \mathcal{L}_{\text{InfoNCE}} = -\frac{1}{2N} \sum_{i=1}^{N}
    \left[
      \log \frac{\exp(S_{ii})}{\sum_j \exp(S_{ij})}
      + \log \frac{\exp(S_{ii})}{\sum_j \exp(S_{ji})}
    \right]
  \label{eq:infonce}
\end{equation}
The first term aligns CSI to text, and the second aligns text to CSI. A key limitation is that the softmax assumes one positive per row or column, which is not fully consistent with our many-to-many setting.

\paragraph{SigLIP}~\citep{zhai2023siglip}.
SigLIP replaces the softmax with independent sigmoid losses over all CSI-text pairs:
\begin{equation}
  \mathcal{L}_{\text{SigLIP}} = -\frac{1}{N^2} \sum_{i,j}
    \log \sigma\!\bigl(t_{ij} \cdot (S_{ij} + b)\bigr)
\end{equation}
where $t_{ij}=+1$ for positive pairs and $-1$ for negative pairs, and $b$ is a learnable bias initialized to $-10.0$. The temperature is initialized to $0.1$. Compared with InfoNCE, this formulation is better suited to cases where one sample may have multiple valid matches.

\paragraph{LanguageBind}~\citep{zhu2024languagebind}.
LanguageBind uses an asymmetric InfoNCE objective:
\begin{equation}
  \mathcal{L}_{\text{LB}} = -\frac{1}{N} \sum_{i=1}^{N}
    \log \frac{\exp(S_{ii})}{\sum_j \exp(S_{ij})}
\end{equation}
Unlike the other methods, it does not learn a text projection head. Instead, it directly matches the CSI representation to the frozen text embedding space. In this way, the text side serves as a fixed semantic target, and only the CSI encoder and CSI projection head are trained.

\paragraph{Barlow Twins}~\citep{zbontar2021barlow}.
Instead of contrasting positive and negative pairs explicitly, Barlow Twins aligns the two modalities by shaping their cross-correlation matrix $\mathcal{R}$ toward the identity matrix:
\begin{equation}
  \mathcal{L}_{\text{BT}} = \underbrace{\sum_i (1 - \mathcal{R}_{ii})^2}_{\text{invariance}}
    + \underbrace{\lambda \sum_i \sum_{j \neq i} \mathcal{R}_{ij}^2}_{\text{redundancy reduction}}
\end{equation}
with $\lambda = 0.005$. The first term encourages corresponding embedding dimensions to agree across modalities, while the second reduces redundancy between different dimensions. Since this loss operates on correlations rather than pairwise similarities, it does not use temperature scaling.

\paragraph{Masked Cross-Modal Prediction.}
This method combines reconstruction and contrastive learning. We first randomly mask 75\% of the CSI time steps, encode the masked signal, and use a 2-layer reconstruction head (Linear $\to$ GELU $\to$ Linear, dimensions $d \to 2d \to d$) to predict the text embedding. The reconstruction loss is
\begin{equation}
  \mathcal{L}_{\text{recon}} = 1 - \frac{1}{N}\sum_{i=1}^{N} \frac{\hat{z}_{\text{text}}^{i\top}\, z_{\text{text}}^i}{\|\hat{z}_{\text{text}}^i\|\;\|z_{\text{text}}^i\|}
\end{equation}
At the same time, we encode the full unmasked CSI and apply the symmetric InfoNCE loss in Eq.~\ref{eq:infonce}. The final objective is
\begin{equation}
  \mathcal{L}_{\text{MCP}} = \mathcal{L}_{\text{recon}} + 0.5 \cdot \mathcal{L}_{\text{InfoNCE}}.
\end{equation}
Intuitively, this objective encourages the model not only to distinguish matched CSI-text pairs, but also to recover text-level semantics from incomplete CSI observations.
 
\section{Text Generation Pipeline Details}
\label{app:textgen}

WiFi CSI signals do not come with natural text annotations. To build CSI--text pairs, we therefore generate descriptions from labels using a two-level text pipeline. Level~1 provides single-attribute captions for all tasks, while Level~2 adds multi-attribute composite captions for multitask evaluation. This design allows the benchmark to support both standard zero-shot recognition and more compositional scene understanding.

\subsection{Level~1: Base Captions}
\label{app:level1}

Level~1 captions describe one attribute at a time, such as an activity class, a gesture, a proximity level, or a user identity label. For each class in each task, we manually write 4--5 English descriptions that express the same semantics using different wording.

\paragraph{Template authoring.}
All templates are hand-written and follow the signal-grounding principle (\S\ref{sec:dataset}): they describe only information that can be inferred from CSI, such as activity type, gesture identity, or proximity. We intentionally exclude deployment-specific details such as device model or room identity, since those details are not part of the sensing target and could introduce dataset-specific shortcuts.

To increase linguistic diversity, each class is associated with several semantically equivalent sentences that vary in phrasing and sentence structure. This gives the model multiple ways to express the same label without relying on automated paraphrasing. Templates are stored as Python lists grouped by dataset and task.

\paragraph{Representative examples.}
\begin{itemize}[leftmargin=*,itemsep=2pt]
\item \textbf{CSI-Bench HAR -- walking:}
  \begin{enumerate}[itemsep=0pt]
    \item ``A person is walking at a normal pace indoors.''
    \item ``Someone walks steadily across the room.''
    \item ``An individual strolls through the monitored area.''
    \item ``A person takes measured steps while walking.''
    \item ``Someone moves forward at a comfortable walking speed.''
  \end{enumerate}
\item \textbf{CSI-Bench HAR -- jumping:}
  ``A person is jumping up and down in place.'' / ``Someone leaps into the air repeatedly.'' / ``An individual performs vertical jumps.''
\item \textbf{XRF55 -- carrying:}
  ``A person carries a heavy object.'' / ``Someone lifts and holds a heavy weight while moving.''
\item \textbf{SignFi -- hello:}
  ``A person signs the word `hello' using hand gestures.''
\end{itemize}

\paragraph{Quality filtering.}
After writing the templates, we apply a simple rule-based filter to remove low-quality entries. Specifically, we discard texts shorter than 10 characters or longer than 300 characters, remove exact duplicates using case-insensitive matching, and exclude entries without alphabetic characters. The remaining 4--5 descriptions per class form the final Level~1 text set.

Each Level~1 description is paired with all CSI samples belonging to that class. In other words, the mapping is many-to-one from text to class, and many-to-many at the sample level once all pairings are constructed.

\paragraph{Dataset-level statistics.}
The number of Level~1 descriptions per class varies slightly by dataset: CSI-Bench tasks use 4--5 descriptions per class, XRF55 uses 4, and SignFi uses 5. Across all 353 classes, this yields 2,291 unique Level~1 descriptions (Table~\ref{tab:dataset-stats}).

\subsection{Level~2: Composite Captions}
\label{app:level2}

While Level~1 captions are sufficient for single-task recognition, they cannot describe a full scene containing multiple attributes at once. To support compositional evaluation, we introduce Level~2 captions for the multitask subset of CSI-Bench, where each of the same 55,684 CSI recordings is labeled with activity, proximity, and user identity simultaneously.

\paragraph{Attribute descriptions.}
We first map each attribute value to a short natural-language phrase:
\begin{itemize}[leftmargin=*,itemsep=1pt]
\item \textbf{Activity:} ``walking at a normal pace,'' ``running quickly,'' ``jumping up and down,'' ``sitting still and breathing calmly,'' ``waving their hand.''
\item \textbf{Proximity:} ``0.5 meters,'' ``1.5 meters,'' ``2.5 meters,'' ``3.5 meters.''
\item \textbf{User identity:} ``User~1,'' ``User~2,'' \ldots, ``User~6,'' ``User~22'' (7 users total).
\end{itemize}

\paragraph{Composition templates.}
We then combine these attribute phrases into full-sentence scene descriptions using four template patterns:
\begin{enumerate}[itemsep=1pt]
\item ``\texttt{\{user\}} is \texttt{\{activity\}} at approximately \texttt{\{distance\}} from the sensor.''
\item ``\texttt{\{user\}} performs \texttt{\{activity\}} while positioned about \texttt{\{distance\}} away from the WiFi receiver.''
\item ``At a distance of \texttt{\{distance\}}, \texttt{\{user\}} is observed \texttt{\{activity\}}.''
\item ``The sensor detects \texttt{\{user\}} \texttt{\{activity\}} roughly \texttt{\{distance\}} from the transmitter.''
\end{enumerate}

\paragraph{Composite statistics.}
The multitask subset contains 5 activity classes, 4 proximity levels, and 7 user identities, giving $5 \times 4 \times 7 = 140$ distinct attribute combinations. Each combination is rendered using the 4 composition templates above, producing four composite descriptions per scene configuration.

These Level~2 captions are added on top of the Level~1 captions rather than replacing them. For example, an HAR sample labeled as ``walking'' still keeps its 5 Level~1 activity descriptions, and also receives 4 composite captions corresponding to its specific user--proximity combination, for a total of 9 texts. The composite captions are shared across the three multitask attributes, while the Level~1 captions remain task-specific.

\paragraph{Role of each level.}
The two levels serve different purposes. Level~1 captions support standard single-task zero-shot classification, where the goal is to recognize one attribute at a time. Level~2 captions enable scene-level reasoning over multiple attributes, which is needed for ambient scene QA (\S\ref{app:ambient-qa}). They also make non-semantic labels such as user identity easier to ground, because identity is described together with interpretable contextual attributes such as activity and distance.

\section{Evaluation Protocol Details}
\label{app:protocols}

This section describes the evaluation protocols used in the main paper. Together, these protocols test three increasingly demanding capabilities: standard zero-shot recognition, generalization to unseen language, and multi-attribute scene understanding.

\subsection{Zero-Shot Classification}
\label{app:zeroshot-details}

We first evaluate whether a trained CSI encoder can recognize classes directly from text prototypes, without learning a task-specific classifier head at test time.

\paragraph{Text prototype construction.}
For each class $c$ in a task with $C$ classes, we collect all Level~1 descriptions $\mathcal{T}_c$ from the text registry. Each description is encoded by the frozen text encoder and then projected by the trained text projection head. We define the class prototype as the $\ell_2$-normalized mean of these text embeddings:
\begin{equation}
  p_c = \frac{1}{|\mathcal{T}_c|} \sum_{t \in \mathcal{T}_c} \frac{h_\omega(g_\phi(t))}{\|h_\omega(g_\phi(t))\|_2}
  \label{eq:prototype}
\end{equation}
where $g_\phi$ denotes the frozen text encoder and $h_\omega$ the learned text projection head.

Averaging over multiple paraphrases makes the prototype less sensitive to any single wording choice and provides a more stable representation of the class semantics.

\paragraph{Classification.}
Given a test CSI sample $x$, we compute its embedding as $z_{\text{csi}} = h_\psi(f_\theta(x))$. We then compare this embedding with all class prototypes using cosine similarity and predict the class with the highest score:
\[
\hat{y} = \arg\max_c \cos(z_{\text{csi}},\, p_c).
\]
We do not apply temperature scaling or post-hoc calibration at inference time; predictions are based directly on raw cosine similarity.

\paragraph{Metrics.}
We report classification accuracy and weighted F1-score on the standard test split of each dataset. All methods are evaluated on the same test sets.

\subsection{Open-Vocabulary Evaluation}
\label{app:open-vocab}

Standard zero-shot evaluation reuses the same template families that appear during training. This leaves open an important question: does the model actually learn language-aligned CSI representations, or does it mainly rely on familiar wording? To test genuine language generalization, we replace the training-time templates with new prompt families that are completely disjoint from them.

\paragraph{Prompt family design.}
We construct eight families of unseen prompts, grouped into three levels of difficulty:

\begin{enumerate}[leftmargin=*,itemsep=2pt]
\item \textbf{Minimal} (1 template per class, without ``A person \ldots'' framing):
  \begin{itemize}[itemsep=1pt]
    \item \textit{Domain expert:} prompts written in clinical or biomechanical language. For example, for \textit{walking}: ``Bipedal ambulation at self-selected comfortable velocity.''
    \item \textit{Layperson:} prompts written in casual everyday language. For example: ``Casually strolling from one side to the other.''
    \item \textit{Sensor-grounded:} prompts written in terms of wireless-signal dynamics. For example: ``Moderate-speed translational displacement with regular bipedal gait periodicity.''
  \end{itemize}

\item \textbf{Anchored} (1 template per class, with a natural sentence frame):
  The domain-expert and sensor-grounded prompts are wrapped in a sentence structure such as ``A person \ldots''. This keeps the vocabulary novel while making the syntax closer to the templates used during training.

\item \textbf{Ensemble} (5 templates per class):
  For each vocabulary style, we write five paraphrases per class and average them into a prototype, mirroring the standard zero-shot setting but with fully unseen text.
\end{enumerate}

\paragraph{Evaluation procedure.}
During evaluation, the CSI encoder and projection heads are kept fixed. For each prompt family, we build new class prototypes by averaging the embeddings of that family's prompts using Eq.~\ref{eq:prototype}. We then run the same cosine-similarity classification procedure as in the standard zero-shot evaluation.

This protocol isolates one question: whether the learned CSI embedding space can align with new language forms that were never seen during training.

\subsection{Ambient Scene QA}
\label{app:ambient-qa}

The previous protocols evaluate recognition of a single attribute at a time. Ambient scene QA is more demanding: the model must answer natural-language questions about different aspects of the same scene, such as who is present, what they are doing, and how far they are from the sensor. This tests whether the learned representation can function as a queryable perception interface rather than just a fixed-label recognizer.

We evaluate this capability at two levels.

\paragraph{Level~1: Per-attribute QA.}
At the first level, we ask one question per attribute and let the model choose among candidate answers.

\begin{table}[ht]
\centering
\small
\caption{Ambient QA question templates and candidate answers.}
\label{tab:qa-templates}
\begin{tabular}{p{2.0cm}p{3.5cm}p{6.5cm}}
\toprule
Attribute & Question & Candidate Answers \\
\midrule
Activity & ``What is the person doing?'' & \{walking, running, jumping, sitting and breathing, waving\} \\
Proximity & ``How far is the person from the sensor?'' & \{``approximately 0.5\,m''; ``approximately 1.5\,m''; ``approximately 2.5\,m''; ``approximately 3.5\,m''\} \\
User identity & ``Who is present?'' & \{``User~1'', ``User~2'', \ldots, ``User~6'', ``User~22''\} \\
\bottomrule
\end{tabular}
\end{table}

For each attribute, the candidate answers are derived from the first Level~1 caption for each class (Table~\ref{tab:qa-templates} shows abbreviated forms). We concatenate each answer with its question in the form ``\texttt{\{question\} \{answer\}}'' and encode the resulting text through the frozen text encoder and learned text projection. The CSI embedding is then compared against all candidate answer embeddings using cosine similarity, and the highest-scoring answer is selected.

We report accuracy separately for each attribute. We also report \textbf{joint accuracy}, which counts a prediction as correct only if all queried attributes are predicted correctly for the same sample.

\paragraph{Level~2: Composite scene QA.}
Level~2 evaluates a more holistic setting. Instead of predicting each attribute independently, the model directly selects from a set of full-scene descriptions. We enumerate all $5 \times 4 \times 7 = 140$ combinations of activity, proximity, and user identity, and encode one composite sentence for each combination using the first composition template:
``\texttt{\{user\}} is \texttt{\{activity\}} at approximately \texttt{\{distance\}} from the sensor.''

For each CSI sample, we compare its embedding against all 140 composite embeddings and select the nearest one by cosine similarity. A prediction is counted as correct only when all three attributes match exactly.

In addition to top-1 accuracy, we report top-3 accuracy, top-5 accuracy, mean reciprocal rank (MRR), and decomposed per-attribute accuracy obtained from the predicted composite description.

Compared with Level~1, this protocol tests whether the model can represent the scene jointly rather than as a product of separate attribute decisions.
 
\section{Text-to-CSI Retrieval}
\label{app:text-to-csi}

Table~\ref{tab:retrieval} in the main text reports CSI-to-text retrieval, where each CSI test embedding is matched against text prototypes. Here we evaluate the reverse direction: given a text prototype, can the model retrieve CSI samples from the corresponding class?

For each class prototype, we rank all CSI test samples by cosine similarity. We then compute Text-to-CSI Recall@$K$ for each class as the fraction of relevant CSI samples appearing in the top-$K$ retrieved results, normalized by $\min(K, n_{\text{relevant}})$, and average this quantity across classes.

This metric should be interpreted slightly differently from standard retrieval settings. Because each class contains many relevant CSI samples, often hundreds or thousands, the metric reflects how cleanly the top-ranked positions are dominated by the correct class, rather than how completely the method covers all relevant samples. For this reason, R@5 or R@10 can occasionally be slightly lower than R@1: even if the top-ranked item is correct, a few incorrect samples entering the top-$K$ list can reduce the normalized score.

\begin{table}[ht]
\centering
\caption{Text-to-CSI retrieval. Left: R@1 (\%) across all tasks. Right: R@5 and R@10 for large-class tasks. \best{Bold}: best per column.}
\label{tab:retrieval-t2c}
\vspace{2pt}
\small
\resizebox{\textwidth}{!}{%
\begin{tabular}{@{}lcccccccccc@{}}
\toprule
 & \multicolumn{7}{c}{\textbf{Text$\to$CSI R@1}} & \multicolumn{2}{c}{\textbf{R@5}} & \multicolumn{1}{c}{\textbf{R@10}} \\
\cmidrule(lr){2-8} \cmidrule(lr){9-10} \cmidrule(l){11-11}
\textbf{Method} & HAR & Proximity & User ID & Fall & MotSrc & SignFi & XRF55 & SignFi & XRF55 & SignFi \\
\midrule
CLIP InfoNCE & \best{100.0} & \best{100.0} & \best{100.0} & \best{100.0} & \best{100.0} & \best{100.0} & 80.0 & \best{99.3} & 82.5 & \best{98.2} \\
SigLIP & \best{100.0} & \best{100.0} & \best{100.0} & \best{100.0} & \best{100.0} & 46.4 & \best{92.7} & 32.0 & \best{93.1} & 35.0 \\
LanguageBind & 80.0 & 25.0 & 83.3 & 50.0 & \best{100.0} & 35.5 & 52.7 & 23.7 & 52.0 & 24.2 \\
Barlow Twins & \best{100.0} & \best{100.0} & \best{100.0} & 0.0 & \best{100.0} & 31.9 & 78.2 & 26.0 & 81.5 & 28.9 \\
Masked Pred. & \best{100.0} & \best{100.0} & \best{100.0} & \best{100.0} & \best{100.0} & 98.2 & 89.1 & 93.8 & 88.7 & 93.6 \\
\bottomrule
\end{tabular}%
}
\end{table}

The results in Table~\ref{tab:retrieval-t2c} largely mirror the main-text retrieval trends, while also revealing some directional differences. CLIP InfoNCE achieves perfect R@1 on 6 of the 7 tasks; the only exception is XRF55, where SigLIP performs best at 92.7\%. On the large-class SignFi task, CLIP InfoNCE remains especially strong, reaching 99.3\% R@5 and 98.2\% R@10, which indicates that the top-ranked CSI samples are almost always dominated by the correct class. Masked Prediction is the strongest alternative on SignFi, with 98.2\% R@1 and 93.8\% R@5.

LanguageBind shows the clearest directional asymmetry. Although it performs reasonably well for CSI-to-text retrieval in the main text, its text-to-CSI R@1 drops substantially on HAR (80.0\%), Fall Detection (50.0\%), and Proximity (25.0\%). This suggests that its asymmetric training objective can organize CSI embeddings well enough to match text prototypes, but does not produce equally tight class-wise CSI clustering.
 
\begin{crblock}
\section{Human-Written Free-Form Language Layer}
\label{app:freeform}

This section gives details of the human-written free-form layer and the comparison in \S\ref{sec:exp-freeform}. Human writers authored descriptions for the five CSI-Bench HAR activities using the collection-session logs and activity descriptions as references, independently of the template registry and without LLM assistance. The descriptions span five registers: conversational, neutral-descriptive, field-note, log-style, and hedged. Representative examples include ``Fast footsteps / someone's running through here,'' ``Log: subject seated, resting respiration, no other motion,'' and ``Someone's waving hello (or maybe goodbye, hard to tell).''

\paragraph{Protocol.}
We train a template-supervised and a free-form-supervised CLIP-InfoNCE model with identical architectures, data splits, and optimization settings (Appendix~\ref{app:training}). Both models are evaluated on the same prompt families with frozen encoders and projection heads, following Appendix~\ref{app:open-vocab}. Held-out free-form prompts are human-written descriptions seen by neither model during training. For ensemble evaluation, five unseen prompt variants per class are averaged into each text prototype (Eq.~\ref{eq:prototype}).

\paragraph{Results.}
Table~\ref{tab:freeform} in \S\ref{sec:exp-freeform} reports the comparison. The template-trained model reaches $95.0\%$ on held-out free-form prompts, $0.9$ points below its training-template accuracy, so its recognition extends beyond the caption formulations used in training. Free-form supervision raises raw domain-expert and sensor-grounded accuracy by $27.9$ and $44.8$ points and the sensor-grounded ensemble by $7.6$ points. It lowers training-template, layperson, and held-out free-form accuracy by $1.4$, $2.1$, and $1.0$ points. Template supervision therefore transfers to ordinary free-form language, and broader human-written supervision improves coverage of specialized language at a small cost on ordinary language. Together with the text-encoder ablation in Table~\ref{tab:ablations}, where replacing CLIP with MiniLM lowers zero-shot accuracy from $98.7\%$ to $90.2\%$, these results are consistent with both training-language coverage and text representation contributing to the gap on specialized prompts.

\section{Many-to-Many Retrieval Metrics}
\label{app:m2m-retrieval}

Recall@$K$ and Rank-1 measure whether a relevant item appears near the top of the ranking. In the many-to-many setting, each CSI sample has several valid descriptions, so we also measure how highly all valid descriptions are ranked. For each CSI test sample, we rank the full gallery of 221 unique HAR captions by cosine similarity and compute Hit@1, average precision, and R-Precision, where $R$ is the number of relevant captions for that query. We report mAP as the mean average precision over queries.

\paragraph{Relevance definitions.}
Under \emph{strict relevance}, the relevant captions are the 9 captions assigned to the queried sample: its 5 Level~1 activity captions and the 4 Level~2 captions of its activity--identity--distance combination. Strict relevance yields 49 distinct relevant sets, one per observed combination. Under \emph{class-level relevance}, the relevant captions are all captions describing the correct activity: its Level~1 captions and the Level~2 captions of every observed combination with that activity. Class-level relevance yields 5 relevant sets. Strict relevance tests whether the model retrieves the descriptions paired with the sample; class-level relevance tests whether it ranks all semantically valid descriptions highly.

\begin{table}[ht]\crcolor
\centering
\caption{CSI$\to$Text retrieval over the full 221-caption HAR gallery (CLIP InfoNCE).}
\label{tab:m2m-retrieval}
\small
\begin{tabular}{@{}lccc@{}}
\toprule
Relevance & Hit@1 & mAP & R-Precision \\
\midrule
Strict & 0.970 & 0.953 & 0.924 \\
Class-level & 0.990 & 0.804 & 0.668 \\
\bottomrule
\end{tabular}
\end{table}

Under class-level relevance, Hit@1 is near saturation at $0.990$, while mAP ($0.804$) and R-Precision ($0.668$) show remaining headroom in ranking all valid descriptions (Table~\ref{tab:m2m-retrieval}).

\section{Held-Out Composition}
\label{app:hoc}

Table~\ref{tab:ambient-qa} shows that composite retrieval outperforms independent per-attribute decisions. This comparison alone does not distinguish composition of separately learned attributes from recall of combinations seen in training. We therefore withhold attribute combinations from training.

\paragraph{Protocol.}
Of the 49 observed activity--identity--distance combinations, we withhold 12, roughly one quarter, such that every constituent attribute value remains present in other training combinations.
Computed over all co-labeled samples (train, validation, and test) before withholding, the normalized mutual information (NMI) between identity and activity is $0.0034$ and between identity and distance is $0.0049$; activity and distance are partially correlated (NMI $0.31$), because walking is the only activity recorded at more than one distance. We evaluate two settings:
\begin{itemize}[leftmargin=*,itemsep=1pt]
\item \textbf{Joint caption withheld:} the CSI recordings and Level~1 captions of the held-out combinations remain in training, and their Level~2 captions are removed.
\item \textbf{Scene withheld:} all recordings and captions of the held-out combinations are removed from training.
\end{itemize}
Both settings are evaluated on the 1,306 test samples of the 12 held-out combinations, using the Level~2 scoring of Appendix~\ref{app:ambient-qa}. Open-set retrieval ranks all 140 composite candidates (chance $0.7\%$); closed-set retrieval ranks only the 12 held-out combinations (chance $8.3\%$). The full-data reference is the model of Table~\ref{tab:ambient-qa}, evaluated on the same 1,306 samples.

\begin{table}[ht]\crcolor
\centering
\caption{Level~2 top-1 accuracy (\%) on the 1,306 test samples of the 12 held-out combinations.}
\label{tab:hoc}
\small
\begin{tabular}{@{}lcc@{}}
\toprule
Training setting & Open-set (140-way) & Closed-set (12-way) \\
\midrule
Full-data reference & 58.5 & 99.2 \\
Joint caption withheld & 49.5 & 89.8 \\
Scene withheld & 13.9 & 63.5 \\
\bottomrule
\end{tabular}
\end{table}

\paragraph{Relation to Table~\ref{tab:ambient-qa}.}
The full-data reference reaches $58.5\%$ on the held-out combinations, below the $86.9\%$ reported over all 4,602 test samples. The two numbers are consistent: 764 of the 1,306 held-out-combination samples and 3,237 of the remaining 3,296 samples ($98.2\%$) are correct, giving 4,001 of 4,602 ($86.9\%$). The held-out combinations are harder than average because 6 of the 12 involve walking, the only activity recorded at multiple distances. Resolving these combinations requires inferring distance, the weakest attribute in Table~\ref{tab:ambient-qa}.

\paragraph{Results.}
With the joint caption withheld, the model retains $84.6\%$ and $90.5\%$ of the full-data accuracy in the open-set and closed-set settings (Table~\ref{tab:hoc}). This is consistent with the model matching a new combination using separately learned activity, identity, and distance information, without that combination's caption appearing in training. With the entire scene withheld, accuracy drops to $13.9\%$ and $63.5\%$, which remains well above chance; unseen signal combinations are harder than unseen caption combinations.

\section{Cross-Environment and Cross-Device Transfer}
\label{app:cross-domain}

The main-text evaluations use matched in-domain splits to isolate the effect of CSI--language alignment. WiFi CSI is sensitive to environment and hardware~\citep{WiFisurveyWang, jiang2018ei, zheng2019widar3}, and deployed systems operate across heterogeneous homes and devices~\citep{zhu2025experience}. We therefore ask whether alignment adds transfer degradation beyond the underlying CSI domain shift.

\paragraph{Protocol.}
We use the cross-environment and cross-device test splits provided with CSI-Bench HAR~\citep{zhu2025csi}. The supervised reference uses the same CSI encoder, source-domain training split, and target-domain evaluation as the aligned model, and is trained with class-label supervision. The aligned model is evaluated zero-shot with text prototypes (Appendix~\ref{app:zeroshot-details}). For target-domain adaptation, we adapt a bottleneck module using $K\in\{1,5,10,20\}$ labeled target-domain samples per class and report mean $\pm$ standard deviation over 5 trials.

\begin{table}[ht]\crcolor
\centering
\caption{HAR accuracy (\%) under cross-environment and cross-device shift. $K$-shot results are mean $\pm$ std over 5 trials.}
\label{tab:cross-domain}
\small
\resizebox{\textwidth}{!}{%
\begin{tabular}{@{}lcccccc@{}}
\toprule
Shift & Supervised & Zero-shot aligned & 1-shot & 5-shot & 10-shot & 20-shot \\
\midrule
Cross-environment & 36.9 & 31.9 & 36.6{\scriptsize$\pm$4.8} & 36.9{\scriptsize$\pm$4.4} & 40.7{\scriptsize$\pm$3.3} & 45.4{\scriptsize$\pm$5.3} \\
Cross-device & 39.6 & 39.9 & 43.1{\scriptsize$\pm$6.0} & 46.7{\scriptsize$\pm$2.3} & 48.8{\scriptsize$\pm$7.0} & 55.0{\scriptsize$\pm$3.5} \\
\bottomrule
\end{tabular}%
}
\end{table}

\paragraph{Results.}
In-domain, the supervised and aligned models reach $97.9\%$ and $98.7\%$. Both degrade under environment and device shift (Table~\ref{tab:cross-domain}). The zero-shot aligned model is $5.0$ points below the supervised reference under cross-environment shift and comparable under cross-device shift ($39.9\%$ vs.\ $39.6\%$), consistent with the CSI domain shift accounting for most of the degradation. Target-domain adaptation improves the aligned model; at 20 shots per class, accuracy reaches $45.4\%$ and $55.0\%$, gains of $13.6$ and $15.1$ points over zero-shot evaluation. Language alignment does not resolve CSI domain shift, and the aligned representation remains adaptable with limited target-domain supervision. We include these splits, the matched supervised reference, and $K$-shot adaptation as benchmark protocols.

\section{Provenance, Intended Uses, and Privacy}
\label{app:datasheet}

\paragraph{Provenance.}
CSI-TextBench is derived from CSI-Bench~\citep{zhu2025csi}, XRF55~\citep{wang2024xrf55}, and SignFi~\citep{ma2019signfi}, each used under its original terms. For SignFi, only preprocessing scripts and CSI--text registry files are released; users obtain the CSI data from the dataset providers.

\paragraph{Intended uses and privacy.}
CSI-TextBench is intended for research on CSI--language alignment, zero-shot and few-shot sensing, and language-queryable ambient perception. The released representation is CSI amplitude and contains no visual or audio recordings, and device and room identifiers are excluded from the captions. CSI can still reveal occupancy and human activity. Out-of-scope uses include covert monitoring, identifying individuals without their consent, and deployment without access control. New data collection or deployment must obtain informed consent and follow applicable institutional and legal requirements.
\end{crblock}

\section{Comparison with Published Source-Dataset Results}
\label{app:sota-context}

To place CSI-TextBench in context, Table~\ref{tab:sota-context} compares our results with the numbers originally reported for each source dataset. This comparison is intended only as a reference point, since training pipelines and evaluation settings are not always identical across papers. Still, it helps show whether our unified framework remains competitive with task-specific baselines.

\begin{table}[ht]
\centering
\caption{CSI-TextBench results in context with published source-dataset numbers. All evaluations are in-domain.}
\label{tab:sota-context}
\vspace{2pt}
\small
\begin{tabular}{@{}llccc@{}}
\toprule
\textbf{Task} & \textbf{Source} & \textbf{Published} & \textbf{Ours (Sup.)} & \textbf{Ours (Zero-Shot)} \\
\midrule
HAR & CSI-Bench~\citep{zhu2025csi} & 75.4 & 97.9 & 97.8 \\
Proximity & CSI-Bench~\citep{zhu2025csi} & 77.5 & 96.3 & 94.6 \\
User ID & CSI-Bench~\citep{zhu2025csi} & 99.5 & 100.0 & 100.0 \\
Fall Detection & CSI-Bench~\citep{zhu2025csi} & 94.9 & 95.5 & 94.2 \\
Motion Source & CSI-Bench~\citep{zhu2025csi} & 99.6 & 99.8 & 99.5 \\
SignFi & SignFi~\citep{ma2019signfi} & 94.8 & 99.4 & 97.0 \\
XRF55 (WiFi) & XRF55~\citep{wang2024xrf55} & ${\sim}$82.9 & 82.9 & 77.5 \\
\bottomrule
\end{tabular}
\end{table}

Overall, the unified CSI-TextBench setup remains highly competitive with previously reported task-specific results. On Fall Detection, Motion Source, and User ID, our supervised baseline is within 0.6 points of the published numbers. On XRF55, our supervised result matches the WiFi accuracy reported by the source paper. On SignFi, both our supervised and zero-shot models exceed the originally reported 94.8\%, indicating that the alignment-based formulation remains effective even for very fine-grained classes.

The largest differences appear on HAR and Proximity, where our supervised baseline is substantially stronger than the numbers reported in CSI-Bench. Importantly, the zero-shot CLIP InfoNCE model remains very close to this supervised baseline, within 0.1--1.7 points across tasks. This supports the main claim of the paper: text-aligned CSI representations can approach standard supervised performance while providing a more flexible interface at inference time.

\section{Embedding Visualization}
\label{app:embedding-viz}

\begin{figure}[ht]
\centering
\includegraphics[width=0.75\linewidth]{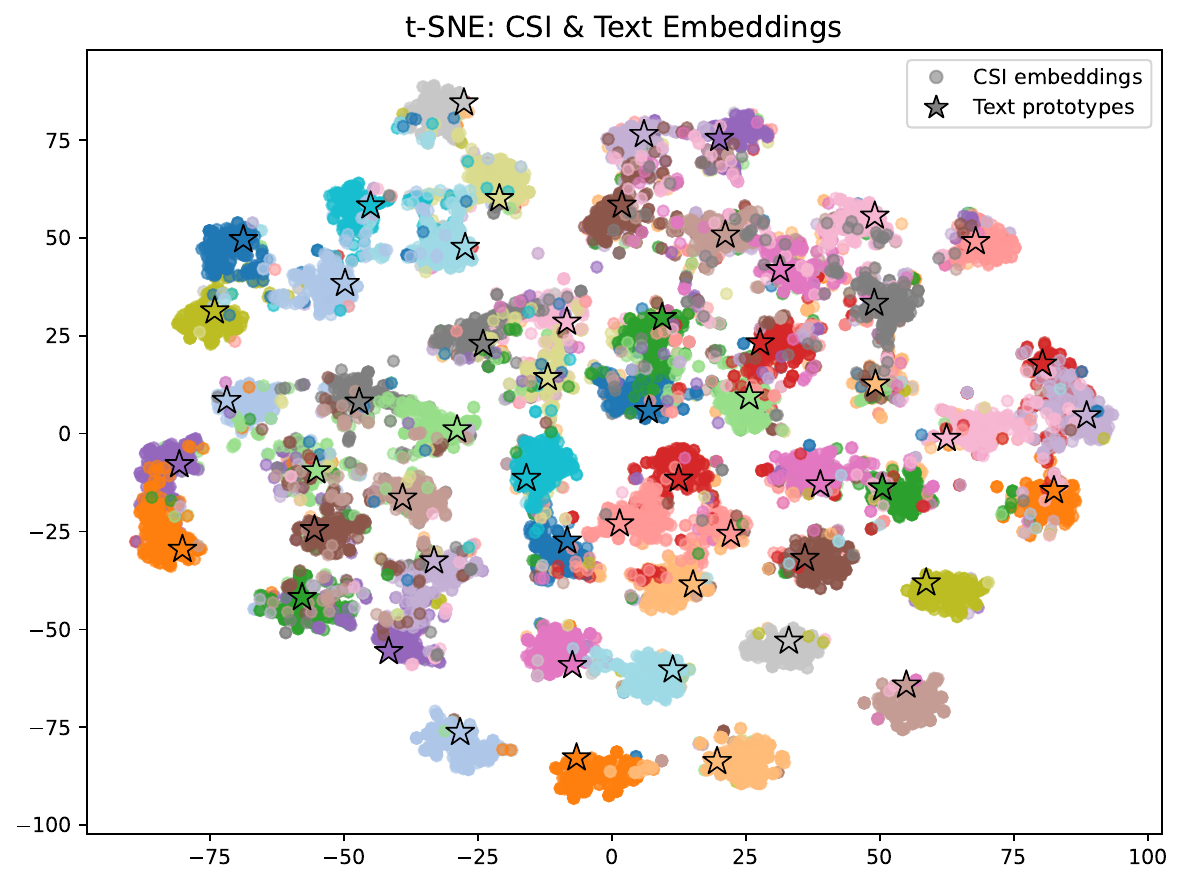}
\caption{t-SNE visualization of aligned CSI and text embeddings for XRF55 action recognition (55 classes, CLIP-InfoNCE). Circles denote CSI sample embeddings; stars denote text prototype embeddings. Matching colors indicate the same action class.}
\label{fig:tsne-xrf55}
\end{figure}

Figure~\ref{fig:tsne-xrf55} provides a qualitative view of the learned CSI--text embedding space on XRF55 action recognition, which contains 55 fine-grained classes. Each circle denotes a CSI test sample encoded by the trained ResNet-18 1D encoder and projection head, and each star denotes the corresponding text prototype obtained by averaging the projected CLIP embeddings of the class paraphrases. Matching colors indicate the same action class.

The main pattern is that text prototypes tend to lie within or close to the corresponding CSI clusters. This suggests that the CLIP-InfoNCE objective successfully brings the two modalities into a shared embedding space, even in a relatively difficult setting with many fine-grained action categories. The figure therefore provides a qualitative complement to the retrieval and zero-shot results reported earlier.

\section{Training Details and Hyperparameters}
\label{app:training}

This section summarizes the shared training setup used across experiments. Unless otherwise noted, all alignment models use the same optimizer, scheduler, projection head, and training protocol.

\begin{itemize}[leftmargin=*,itemsep=2pt]
\item \textbf{Optimizer:} AdamW with learning rate $1\!\times\!10^{-3}$ and weight decay $1\!\times\!10^{-5}$.
\item \textbf{Scheduler:} Cosine annealing with 5-epoch linear warmup.
\item \textbf{Gradient clipping:} Maximum gradient norm of $1.0$.
\item \textbf{Batch size:} 128 for both single-task and multitask training.
\item \textbf{Mixed precision:} bfloat16 automatic mixed precision.
\item \textbf{Training length:} Up to 200 epochs, with early stopping based on validation loss using patience 15.
\item \textbf{Temperature:} Learnable $\tau = 0.07$ stored in log-space for all methods except SigLIP, which uses $\tau = 0.1$ together with a learnable bias $b = -10.0$.
\item \textbf{Projection head:} A 2-layer MLP of size input\_dim $\to$ 1024 $\to$ 256, with BatchNorm and GELU after the hidden layer. The output is $\ell_2$-normalized. We initialize the weights with truncated normal noise ($\sigma{=}0.02$).
\item \textbf{Supervised baseline:} The same ResNet-18 1D encoder followed by a linear classification head, trained with cross-entropy and label smoothing 0.1.
\item \textbf{Hardware:} AWS SageMaker ml.g5.xlarge with one NVIDIA A10G GPU (24\,GB).
\item \textbf{Training time:} Approximately 15--45 minutes per single-task run and 2--4 hours for multitask-all training.
\item \textbf{Random seed:} 42 for all experiments. 123, 456 are also used for std calculation. 
\end{itemize}
 
 
 
\newpage
\section*{NeurIPS Paper Checklist}

\begin{enumerate}

\item {\bf Claims}
    \item[] Question: Do the main claims made in the abstract and introduction accurately reflect the paper's contributions and scope?
    \item[] Answer: \answerYes{}
    \item[] Justification: The abstract and Section~\ref{sec:intro} state three contributions (large-scale CSI--language dataset, controlled alignment benchmarking, and composable ambient perception) that are supported by the dataset construction, method formulation, and experiments in Sections~\ref{sec:dataset}--\ref{sec:experiments}.
    \item[] Guidelines:
    \begin{itemize}
        \item The answer \answerNA{} means that the abstract and introduction do not include the claims made in the paper.
        \item The abstract and/or introduction should clearly state the claims made, including the contributions made in the paper and important assumptions and limitations. A \answerNo{} or \answerNA{} answer to this question will not be perceived well by the reviewers. 
        \item The claims made should match theoretical and experimental results, and reflect how much the results can be expected to generalize to other settings. 
        \item It is fine to include aspirational goals as motivation as long as it is clear that these goals are not attained by the paper. 
    \end{itemize}

\item {\bf Limitations}
    \item[] Question: Does the paper discuss the limitations of the work performed by the authors?
    \item[] Answer: \answerYes{} 
    \item[] Justification: \cradd{Section~\ref{sec:limitations} discusses the limitations, including the scope of open-vocabulary evaluation and cross-domain transfer (Appendix~\ref{app:cross-domain}).} 
    \item[] Guidelines:
    \begin{itemize}
        \item The answer \answerNA{} means that the paper has no limitation while the answer \answerNo{} means that the paper has limitations, but those are not discussed in the paper. 
        \item The authors are encouraged to create a separate ``Limitations'' section in their paper.
        \item The paper should point out any strong assumptions and how robust the results are to violations of these assumptions (e.g., independence assumptions, noiseless settings, model well-specification, asymptotic approximations only holding locally). The authors should reflect on how these assumptions might be violated in practice and what the implications would be.
        \item The authors should reflect on the scope of the claims made, e.g., if the approach was only tested on a few datasets or with a few runs. In general, empirical results often depend on implicit assumptions, which should be articulated.
        \item The authors should reflect on the factors that influence the performance of the approach. For example, a facial recognition algorithm may perform poorly when image resolution is low or images are taken in low lighting. Or a speech-to-text system might not be used reliably to provide closed captions for online lectures because it fails to handle technical jargon.
        \item The authors should discuss the computational efficiency of the proposed algorithms and how they scale with dataset size.
        \item If applicable, the authors should discuss possible limitations of their approach to address problems of privacy and fairness.
        \item While the authors might fear that complete honesty about limitations might be used by reviewers as grounds for rejection, a worse outcome might be that reviewers discover limitations that aren't acknowledged in the paper. The authors should use their best judgment and recognize that individual actions in favor of transparency play an important role in developing norms that preserve the integrity of the community. Reviewers will be specifically instructed to not penalize honesty concerning limitations.
    \end{itemize}

\item {\bf Theory assumptions and proofs}
    \item[] Question: For each theoretical result, does the paper provide the full set of assumptions and a complete (and correct) proof?
    \item[] Answer: \answerNA{}
    \item[] Justification: The paper does not include theoretical results such as theorems or proofs. The alignment loss formulations in Section~\ref{sec:methods} and Appendix~\ref{app:methods} are standard objectives from prior work and are cited accordingly.
    \item[] Guidelines:
    \begin{itemize}
        \item The answer \answerNA{} means that the paper does not include theoretical results. 
        \item All the theorems, formulas, and proofs in the paper should be numbered and cross-referenced.
        \item All assumptions should be clearly stated or referenced in the statement of any theorems.
        \item The proofs can either appear in the main paper or the supplemental material, but if they appear in the supplemental material, the authors are encouraged to provide a short proof sketch to provide intuition. 
        \item Inversely, any informal proof provided in the core of the paper should be complemented by formal proofs provided in appendix or supplemental material.
        \item Theorems and Lemmas that the proof relies upon should be properly referenced. 
    \end{itemize}

    \item {\bf Experimental result reproducibility}
    \item[] Question: Does the paper fully disclose all the information needed to reproduce the main experimental results of the paper to the extent that it affects the main claims and/or conclusions of the paper (regardless of whether the code and data are provided or not)?
    \item[] Answer: \answerYes{}
    \item[] Justification: The paper provides full architecture details (Section~\ref{sec:exp-setup}), all hyperparameters (Appendix~\ref{app:training}), preprocessing pipeline (Appendix~\ref{app:preprocessing}), alignment loss formulations (Appendix~\ref{app:methods}), evaluation protocols (Appendix~\ref{app:protocols}), \cradd{text generation details (Appendix~\ref{app:textgen}), and the protocols for free-form language, many-to-many retrieval, held-out composition, and cross-domain transfer (Appendices~\ref{app:freeform}--\ref{app:cross-domain}).} The dataset, code, and benchmark are released.
    \item[] Guidelines:
    \begin{itemize}
        \item The answer \answerNA{} means that the paper does not include experiments.
        \item If the paper includes experiments, a \answerNo{} answer to this question will not be perceived well by the reviewers: Making the paper reproducible is important, regardless of whether the code and data are provided or not.
        \item If the contribution is a dataset and\slash or model, the authors should describe the steps taken to make their results reproducible or verifiable. 
        \item Depending on the contribution, reproducibility can be accomplished in various ways. For example, if the contribution is a novel architecture, describing the architecture fully might suffice, or if the contribution is a specific model and empirical evaluation, it may be necessary to either make it possible for others to replicate the model with the same dataset, or provide access to the model. In general. releasing code and data is often one good way to accomplish this, but reproducibility can also be provided via detailed instructions for how to replicate the results, access to a hosted model (e.g., in the case of a large language model), releasing of a model checkpoint, or other means that are appropriate to the research performed.
        \item While NeurIPS does not require releasing code, the conference does require all submissions to provide some reasonable avenue for reproducibility, which may depend on the nature of the contribution. For example
        \begin{enumerate}
            \item If the contribution is primarily a new algorithm, the paper should make it clear how to reproduce that algorithm.
            \item If the contribution is primarily a new model architecture, the paper should describe the architecture clearly and fully.
            \item If the contribution is a new model (e.g., a large language model), then there should either be a way to access this model for reproducing the results or a way to reproduce the model (e.g., with an open-source dataset or instructions for how to construct the dataset).
            \item We recognize that reproducibility may be tricky in some cases, in which case authors are welcome to describe the particular way they provide for reproducibility. In the case of closed-source models, it may be that access to the model is limited in some way (e.g., to registered users), but it should be possible for other researchers to have some path to reproducing or verifying the results.
        \end{enumerate}
    \end{itemize}

\item {\bf Open access to data and code}
    \item[] Question: Does the paper provide open access to the data and code, with sufficient instructions to faithfully reproduce the main experimental results, as described in supplemental material?
    \item[] Answer: \answerYes{}
    \item[] Justification: The paper states the release of the dataset, code, and benchmark in the abstract and reiterates it in Section~\ref{sec:conclusion}. A code repository is provided with the submission.
    \item[] Guidelines:
    \begin{itemize}
        \item The answer \answerNA{} means that paper does not include experiments requiring code.
        \item Please see the NeurIPS code and data submission guidelines (\url{https://neurips.cc/public/guides/CodeSubmissionPolicy}) for more details.
        \item While we encourage the release of code and data, we understand that this might not be possible, so \answerNo{} is an acceptable answer. Papers cannot be rejected simply for not including code, unless this is central to the contribution (e.g., for a new open-source benchmark).
        \item The instructions should contain the exact command and environment needed to run to reproduce the results. See the NeurIPS code and data submission guidelines (\url{https://neurips.cc/public/guides/CodeSubmissionPolicy}) for more details.
        \item The authors should provide instructions on data access and preparation, including how to access the raw data, preprocessed data, intermediate data, and generated data, etc.
        \item The authors should provide scripts to reproduce all experimental results for the new proposed method and baselines. If only a subset of experiments are reproducible, they should state which ones are omitted from the script and why.
        \item At submission time, to preserve anonymity, the authors should release anonymized versions (if applicable).
        \item Providing as much information as possible in supplemental material (appended to the paper) is recommended, but including URLs to data and code is permitted.
    \end{itemize}

\item {\bf Experimental setting/details}
    \item[] Question: Does the paper specify all the training and test details (e.g., data splits, hyperparameters, how they were chosen, type of optimizer) necessary to understand the results?
    \item[] Answer: \answerYes{}
    \item[] Justification: Training details including optimizer, learning rate, batch size, scheduler, and hardware are in Section~\ref{sec:exp-setup} and Appendix~\ref{app:training}. Evaluation protocols are detailed in Appendix~\ref{app:protocols} \cradd{and Appendices~\ref{app:freeform}--\ref{app:cross-domain}}. Data splits follow the source datasets' standard protocols\cradd{, including the CSI-Bench cross-environment and cross-device splits}.
    \item[] Guidelines:
    \begin{itemize}
        \item The answer \answerNA{} means that the paper does not include experiments.
        \item The experimental setting should be presented in the core of the paper to a level of detail that is necessary to appreciate the results and make sense of them.
        \item The full details can be provided either with the code, in appendix, or as supplemental material.
    \end{itemize}

\item {\bf Experiment statistical significance}
    \item[] Question: Does the paper report error bars suitably and correctly defined or other appropriate information about the statistical significance of the experiments?
    \item[] Answer: \answerYes{}
    \item[] Justification: Table~\ref{tab:zeroshot} reports mean $\pm$ standard deviation over 3 random seeds for zero-shot classification. Few-shot results in Table~\ref{tab:holdout} report mean $\pm$ std over 10 trials\cradd{, and target-domain adaptation results in Table~\ref{tab:cross-domain} report mean $\pm$ std over 5 trials}. The source of variability (random seed or trial sampling) is stated in each case.
    \item[] Guidelines:
    \begin{itemize}
        \item The answer \answerNA{} means that the paper does not include experiments.
        \item The authors should answer \answerYes{} if the results are accompanied by error bars, confidence intervals, or statistical significance tests, at least for the experiments that support the main claims of the paper.
        \item The factors of variability that the error bars are capturing should be clearly stated (for example, train/test split, initialization, random drawing of some parameter, or overall run with given experimental conditions).
        \item The method for calculating the error bars should be explained (closed form formula, call to a library function, bootstrap, etc.)
        \item The assumptions made should be given (e.g., Normally distributed errors).
        \item It should be clear whether the error bar is the standard deviation or the standard error of the mean.
        \item It is OK to report 1-sigma error bars, but one should state it. The authors should preferably report a 2-sigma error bar than state that they have a 96\% CI, if the hypothesis of Normality of errors is not verified.
        \item For asymmetric distributions, the authors should be careful not to show in tables or figures symmetric error bars that would yield results that are out of range (e.g., negative error rates).
        \item If error bars are reported in tables or plots, the authors should explain in the text how they were calculated and reference the corresponding figures or tables in the text.
    \end{itemize}

\item {\bf Experiments compute resources}
    \item[] Question: For each experiment, does the paper provide sufficient information on the computer resources (type of compute workers, memory, time of execution) needed to reproduce the experiments?
    \item[] Answer: \answerYes{}
    \item[] Justification: Section~\ref{sec:exp-setup} and Appendix~\ref{app:training} report the hardware (AWS SageMaker ml.g5.xlarge with NVIDIA A10G, 24\,GB), training time per run (15--45\,min single-task, 2--4\,h multitask), and mixed-precision settings.
    \item[] Guidelines:
    \begin{itemize}
        \item The answer \answerNA{} means that the paper does not include experiments.
        \item The paper should indicate the type of compute workers CPU or GPU, internal cluster, or cloud provider, including relevant memory and storage.
        \item The paper should provide the amount of compute required for each of the individual experimental runs as well as estimate the total compute. 
        \item The paper should disclose whether the full research project required more compute than the experiments reported in the paper (e.g., preliminary or failed experiments that didn't make it into the paper). 
    \end{itemize}
    
\item {\bf Code of ethics}
    \item[] Question: Does the research conducted in the paper conform, in every respect, with the NeurIPS Code of Ethics \url{https://neurips.cc/public/EthicsGuidelines}?
    \item[] Answer: \answerYes{}
    \item[] Justification: The research conforms with the NeurIPS Code of Ethics. The paper uses publicly available datasets collected under their respective IRB approvals (Section~\ref{sec:broader-impact}) and discusses privacy considerations.
    \item[] Guidelines:
    \begin{itemize}
        \item The answer \answerNA{} means that the authors have not reviewed the NeurIPS Code of Ethics.
        \item If the authors answer \answerNo, they should explain the special circumstances that require a deviation from the Code of Ethics.
        \item The authors should make sure to preserve anonymity (e.g., if there is a special consideration due to laws or regulations in their jurisdiction).
    \end{itemize}

\item {\bf Broader impacts}
    \item[] Question: Does the paper discuss both potential positive societal impacts and negative societal impacts of the work performed?
    \item[] Answer: \answerYes{}
    \item[] Justification: Section~\ref{sec:broader-impact} discusses both positive impacts (assistive applications, privacy advantages over camera/microphone-based sensing) and potential negative impacts (activity detection and presence monitoring), along with mitigation strategies such as explicit consent mechanisms and on-device processing.
    \item[] Guidelines:
    \begin{itemize}
        \item The answer \answerNA{} means that there is no societal impact of the work performed.
        \item If the authors answer \answerNA{} or \answerNo, they should explain why their work has no societal impact or why the paper does not address societal impact.
        \item Examples of negative societal impacts include potential malicious or unintended uses (e.g., disinformation, generating fake profiles, surveillance), fairness considerations (e.g., deployment of technologies that could make decisions that unfairly impact specific groups), privacy considerations, and security considerations.
        \item The conference expects that many papers will be foundational research and not tied to particular applications, let alone deployments. However, if there is a direct path to any negative applications, the authors should point it out. For example, it is legitimate to point out that an improvement in the quality of generative models could be used to generate Deepfakes for disinformation. On the other hand, it is not needed to point out that a generic algorithm for optimizing neural networks could enable people to train models that generate Deepfakes faster.
        \item The authors should consider possible harms that could arise when the technology is being used as intended and functioning correctly, harms that could arise when the technology is being used as intended but gives incorrect results, and harms following from (intentional or unintentional) misuse of the technology.
        \item If there are negative societal impacts, the authors could also discuss possible mitigation strategies (e.g., gated release of models, providing defenses in addition to attacks, mechanisms for monitoring misuse, mechanisms to monitor how a system learns from feedback over time, improving the efficiency and accessibility of ML).
    \end{itemize}
    
\item {\bf Safeguards}
    \item[] Question: Does the paper describe safeguards that have been put in place for responsible release of data or models that have a high risk for misuse (e.g., pre-trained language models, image generators, or scraped datasets)?
    \item[] Answer: \answerNA{}
    \item[] Justification: The released dataset contains WiFi CSI amplitude data and templated text descriptions, which do not pose direct risks for misuse comparable to pretrained language models or image generators. The underlying source datasets were already publicly available.
    \item[] Guidelines:
    \begin{itemize}
        \item The answer \answerNA{} means that the paper poses no such risks.
        \item Released models that have a high risk for misuse or dual-use should be released with necessary safeguards to allow for controlled use of the model, for example by requiring that users adhere to usage guidelines or restrictions to access the model or implementing safety filters. 
        \item Datasets that have been scraped from the Internet could pose safety risks. The authors should describe how they avoided releasing unsafe images.
        \item We recognize that providing effective safeguards is challenging, and many papers do not require this, but we encourage authors to take this into account and make a best faith effort.
    \end{itemize}

\item {\bf Licenses for existing assets}
    \item[] Question: Are the creators or original owners of assets (e.g., code, data, models), used in the paper, properly credited and are the license and terms of use explicitly mentioned and properly respected?
    \item[] Answer: \answerYes{}
    \item[] Justification: All three source datasets (CSI-Bench, XRF55, SignFi) are cited with their original papers in Section~\ref{sec:dataset}. The pretrained CLIP model is also cited. The terms of use of the dataset are properly respected.
    \item[] Guidelines:
    \begin{itemize}
        \item The answer \answerNA{} means that the paper does not use existing assets.
        \item The authors should cite the original paper that produced the code package or dataset.
        \item The authors should state which version of the asset is used and, if possible, include a URL.
        \item The name of the license (e.g., CC-BY 4.0) should be included for each asset.
        \item For scraped data from a particular source (e.g., website), the copyright and terms of service of that source should be provided.
        \item If assets are released, the license, copyright information, and terms of use in the package should be provided. For popular datasets, \url{paperswithcode.com/datasets} has curated licenses for some datasets. Their licensing guide can help determine the license of a dataset.
        \item For existing datasets that are re-packaged, both the original license and the license of the derived asset (if it has changed) should be provided.
        \item If this information is not available online, the authors are encouraged to reach out to the asset's creators.
    \end{itemize}

\item {\bf New assets}
    \item[] Question: Are new assets introduced in the paper well documented and is the documentation provided alongside the assets?
   \item[] Answer: \answerYes{}
    \item[] Justification: CSI-TextBench is a new dataset and benchmark. The paper documents the dataset construction (Section~\ref{sec:dataset}, Appendix~\ref{app:textgen}), statistics (Table~\ref{tab:dataset-stats}), preprocessing pipeline (Appendix~\ref{app:preprocessing}), \cradd{evaluation protocols (Appendix~\ref{app:protocols}), and intended uses and privacy considerations (Appendix~\ref{app:datasheet}).} The access to dataset, code, and benchmark is included in the paper.
    \item[] Guidelines:
    \begin{itemize}
        \item The answer \answerNA{} means that the paper does not release new assets.
        \item Researchers should communicate the details of the dataset\slash code\slash model as part of their submissions via structured templates. This includes details about training, license, limitations, etc. 
        \item The paper should discuss whether and how consent was obtained from people whose asset is used.
        \item At submission time, remember to anonymize your assets (if applicable). You can either create an anonymized URL or include an anonymized zip file.
    \end{itemize}

\item {\bf Crowdsourcing and research with human subjects}
    \item[] Question: For crowdsourcing experiments and research with human subjects, does the paper include the full text of instructions given to participants and screenshots, if applicable, as well as details about compensation (if any)? 
    \item[] Answer: \answerNA{}
    \item[] Justification: The paper does not involve crowdsourcing or new research with human subjects. All \cradd{CSI} data comes from existing publicly available datasets.
    \item[] Guidelines:
    \begin{itemize}
        \item The answer \answerNA{} means that the paper does not involve crowdsourcing nor research with human subjects.
        \item Including this information in the supplemental material is fine, but if the main contribution of the paper involves human subjects, then as much detail as possible should be included in the main paper. 
        \item According to the NeurIPS Code of Ethics, workers involved in data collection, curation, or other labor should be paid at least the minimum wage in the country of the data collector. 
    \end{itemize}

\item {\bf Institutional review board (IRB) approvals or equivalent for research with human subjects}
    \item[] Question: Does the paper describe potential risks incurred by study participants, whether such risks were disclosed to the subjects, and whether Institutional Review Board (IRB) approvals (or an equivalent approval/review based on the requirements of your country or institution) were obtained?
    \item[] Answer: \answerNA{}
    \item[] Justification: The paper does not conduct new human subjects research. It builds on publicly available datasets collected under their respective IRB approvals, as noted in Section~\ref{sec:broader-impact}.
    \item[] Guidelines:
    \begin{itemize}
        \item The answer \answerNA{} means that the paper does not involve crowdsourcing nor research with human subjects.
        \item Depending on the country in which research is conducted, IRB approval (or equivalent) may be required for any human subjects research. If you obtained IRB approval, you should clearly state this in the paper. 
        \item We recognize that the procedures for this may vary significantly between institutions and locations, and we expect authors to adhere to the NeurIPS Code of Ethics and the guidelines for their institution. 
        \item For initial submissions, do not include any information that would break anonymity (if applicable), such as the institution conducting the review.
    \end{itemize}

\item {\bf Declaration of LLM usage}
    \item[] Question: Does the paper describe the usage of LLMs if it is an important, original, or non-standard component of the core methods in this research? Note that if the LLM is used only for writing, editing, or formatting purposes and does \emph{not} impact the core methodology, scientific rigor, or originality of the research, declaration is not required.
    \item[] Answer: \answerNA{}
    \item[] Justification: The core methodology does not involve LLMs. \cradd{Template descriptions are hand-written (Appendix~\ref{app:textgen}), and the free-form HAR descriptions are human-written without LLM assistance (Appendix~\ref{app:freeform}).} A frozen CLIP text encoder is used as a feature extractor, which is a standard pretrained component rather than an LLM.
    \item[] Guidelines:
    \begin{itemize}
        \item The answer \answerNA{} means that the core method development in this research does not involve LLMs as any important, original, or non-standard components.
        \item Please refer to our LLM policy in the NeurIPS handbook for what should or should not be described.
    \end{itemize}

\end{enumerate}

\end{document}